\documentclass[sigconf]{acmart}
\renewcommand\footnotetextcopyrightpermission[1]{} 
\usepackage[T1]{fontenc}    
\usepackage{hyperref}       
\usepackage{url}            
\usepackage{booktabs}       
\usepackage{amsfonts}       
\usepackage{nicefrac}       
\usepackage{microtype}      
\usepackage{xcolor}         
\usepackage{wrapfig}
\usepackage{graphicx}
\usepackage{amsthm,amsmath,amsfonts}
\usepackage{subfigure}
\usepackage{color}
\usepackage{bm}
\usepackage{multirow} 
\usepackage{array}
\usepackage{algorithm} 
\usepackage{algorithmicx}
\usepackage{algpseudocode}
\usepackage{booktabs}

\usepackage{tcolorbox}

\def\bH{\textbf{H}}

\newcolumntype{L}[1]{>{\raggedright\let\newline\\\arraybackslash\hspace{0pt}}m{#1}}
\newcolumntype{C}[1]{>{\centering\let\newline  \\\arraybackslash\hspace{0pt}}m{#1}}
\newcolumntype{R}[1]{>{\raggedleft\let\newline \\\arraybackslash\hspace{0pt}}m{#1}}

\usepackage{multirow}
\usepackage[normalem]{ulem}
\useunder{\uline}{\ul}{}

\usepackage{colortbl}

\newcommand\gray{\cellcolor[rgb]{.8,.8,.8}}

\begin{document}
\title{A Versatile Graph Learning Approach through LLM-based Agent}

\author{Lanning Wei$^{1,2}$, 
	Huan Zhao$^{3, \ast}$, 
	Xiaohan Zheng$^{4}$
	Zhiqiang He$^{1,5}$,
	Quanming Yao$^{4,\ast}$}
\affiliation{
	\institution{$^1$Institute of Computing Technology, Chinese Academy of Sciences $^2$University of Chinese Academy of Sciences $^3$Noumena AI, $^4$Department of Electronic Engineering, Tsinghua University, $^5$Lenovo}
	\city{Beijing}
	\country{China}
}
\email{weilanning@163.com; zhaohuan@noumena.com.cn; qyaoaa@tsinghua.edu.cn}

\begin{abstract}

Designing versatile graph learning approaches is important, considering the diverse graphs and tasks existing in real-world applications.
Existing methods have attempted to achieve this target through automated machine learning techniques, pre-training and fine-tuning strategies, and large language models.
However, these methods are not versatile enough for graph learning, as they work on either limited types of graphs or a single task.
In this paper, we propose to explore versatile graph learning approaches with LLM-based agents,
and the key insight is customizing the graph learning procedures for diverse graphs and tasks.
To achieve this, we develop several LLM-based agents, equipped with diverse profiles, tools, functions and human experience.
They collaborate to configure each procedure with task and data-specific settings step by step towards versatile solutions, and the proposed method is dubbed GL-Agent.
By evaluating on diverse tasks and graphs, the correct results of the agent and its comparable performance showcase the versatility of the proposed method, especially in complex scenarios.The low resource cost and the potential to use open-source LLMs highlight the efficiency of GL-Agent.

\end{abstract}

\maketitle

\section{Introduction}
\label{sec-intro}
Graph-structured data have been widely employed in various real-world domains, such as social networks \cite{hamilton2017inductive}, e-commerce graphs \cite{lu2012recommender}, and chemistry or biomedical molecules \cite{zhang2023artificial, gilmer2017neural}. 
These applications, which are based on graphs, display a significant amount of diversity in terms of the domain knowledge and learning tasks contained in the graph. 
These diversities on graphs and learning tasks demands different settings in graph learning procedures toward effective graph mining~\cite{you2020design,rossi2020sign,gao2019graphnas,yao2018taking}.

Existing methods are not versatile enough on graph learning.
To be specific, automated machine learning (AutoML) techniques have been explored towards handling diverse graphs with a single task~\cite{luo2019autocross,wang2022autofield,gao2019graphnas,you2020design,zhang2021automated}. 
It is achieved by customizing the machine learning pipeline in a data-driven manner,
which alleviate the heavy efforts when facing diverse graphs.
To handling diverse graph learning tasks, pre-training and fine-tuning paradigm~\cite{lu2021learning}, especially the prompt tuning strategy from large language models~(LLMs)~\cite{sun2022gppt,liu2023graphprompt}, 
have been widely explored in recent years.
The pre-trained model can adapt to the diverse downstream tasks by tuning with task-specific data.
Furthermore, motivated by the 
emergent ability in LLMs~\cite{brown2020language,touvron2023llama,chowdhery2022palm}, 
flatten-based methods transform graph learning problems into natural language question-answer problems, 
and then use LLMs to obtain the
predictions~\cite{fatemi2023talk,li2023survey}. 
These methods have difficulties in describing large-scale graphs considering the limited context length of LLMs. 

\begin{figure*}[t]
	\centering
	\includegraphics[width=0.9\linewidth]{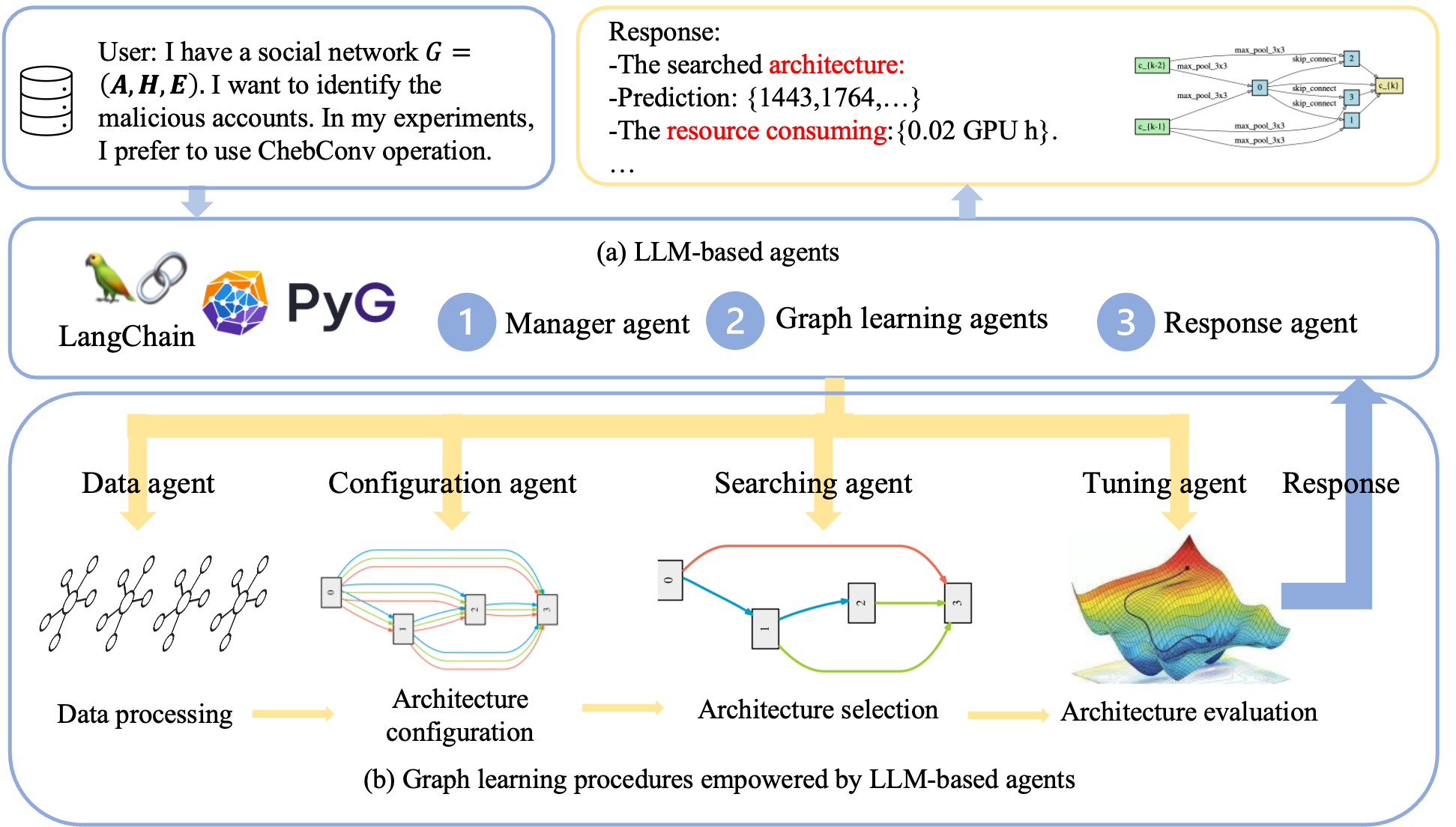}
	\caption{The framework designed to customize the graph learning procedures with LLM-based agents.}
	\label{fig-outline}
\end{figure*}

In this paper, 
we propose GL-Agent, 
which leverage the insight from the versatility of LLMs,
to explore the versatile graph learning approaches.
Learning on diverse tasks and graphs requires different settings in machine learning pipeline, as shown in Fig.~\ref{fig-outline}, i.e., designing different strategies for data, model and hyper-parameters based on the human expertise.
It represents that versatile solutions can be obtained by configuring the pipeline with task- and data-specific settings.
However, it is difficult to achieve this by LLMs directly due to the complex procedures~\cite{wei2022chain,valmeekam2024planbench}.
Considering this, 
we design LLM-based agent in each procedure to configure and complete the pipeline step by step, and the proposed method is dubbed GL-Agent (\uline{G}raph \uline{L}earning Agent).

As shown in Fig.~\ref{fig-outline},  
started with users' instructions of graph learning requirements, 
the manager agent extracts the graph learning keywords from the instructions, e.g., data, learning target, model design preferences.
Then, in graph learning procedures, 
AutoML is adopted due to its effectiveness in designing data-specific approaches ~\cite{yao2018taking,pham2018efficient,gao2019graphnas}.
Four graph learning agents are provided, each specializing for data, AutoML configuration, data-specific model selection, and hyper-parameter optimization, respectively. 
The final response is generated based on the graph learning results, after which the user obtains the response according to the given instructions.
The agent is equipped with well-defined profile, external tools, functions that may used, and expertise description when executing the given procedure. Then, LLMs are able to configure and complete the procedure step by step.

To validate the proposed GL-Agent, we adopt 11 widely used datasets from node, link and graph levels. The versatility and effectiveness are demonstrated by the correct agent outputs and comparable model performance on these datasets, particularly with complex non-homophilous graphs and complex instructions.
The low time and economic cost involved in designing versatile graph learning solution represent the efficiency of GL-Agent.

The main contributions are summarized as follows: 
1) We propose a method GL-Agent to explore the versatile graph learning approaches with LLM-based agent. 
2) Given diverse tasks and graphs, we develop well-equipped manager, graph learning and response agents to configure and conduct the graph learning procedures step by step towards versatile graph learning solutions.
3) Extensive experiments have been conducted on diverse tasks and graphs. The comparable performance and correct agent outputs demonstrate the versatility and effectiveness of the proposed GL-Agent. The low resource cost and potential for using open-source LLMs highlight the efficiency of the proposed method.


\section{Related Work}

\subsection{Versatile Graph Learning Methods}
\label{sec-related-autograph}
Existing methods have widely explored in handling the diverse tasks and data in graph learning problems.

1) \textbf{AutoML} is the representative technique in handling different data by designing data-specific solutions~\cite{yao2018taking}.
Existing methods have explored the data-specific feature selection strategies~\cite{wang2022autofield,luo2019autocross}, feature dimension design methods~\cite{ginart2021mixed,zhaok2021autoemb,liu2020automated}. GNNs are automatically designed from the aspects of operations~\cite{gao2019graph,zhao2020simplifying,zhao2021search,wei2021pooling,you2020design,wang2022profiling} and skip-connections~\cite{wei2022designing,li2020autograph}.
These methods have the ability in handing different graphs while limited in single task as represented in Table~\ref{tb-task-data-comp}.
\begin{table}
	\centering
	\small
	\caption{The comparisons of existing methods in handling diverse tasks and graphs.}
	\begin{tabular}{l|ll}
		\toprule
		Methods & Diverse Tasks                   & Diverse graphs     \\ \midrule
		GL-Agent                            & $\checkmark$                      & $\checkmark$         \\  \midrule
		AutoML             & $\times$                        & $\checkmark$         \\  \midrule
		Pre-training and tuning & $\checkmark$ & $\times$         \\  \midrule
		Flatten-based methods      & $\checkmark$                      & $\checkmark$(small-scale) \\ \bottomrule
	\end{tabular}
	\label{tb-task-data-comp}
\end{table}
2) \textbf{Pre-training and fine-tuning} paradigm, especially prompt tuning, are widely used in recent years to handling the different graph learning tasks, motivated by the versatility of LLMs~\cite{liu2023towards,lu2021learning,liu2023pre,xiong2024large}.
These methods learn the task-specific vector in tuning stage with downstream task data, and then the pre-trained models could applied to different graph learning tasks~\cite{liu2023graphprompt,sun2022gppt,sun2023all,yu2023multigprompt}.
However, the gap between two stages are significant, which representing the deficiency in handling diverse graphs.
3) \textbf{Flatten-based} graph prediction methods are proposed motivated by the emerging ability of LLMs~\cite{fatemi2023talk,wang2024can,li2023survey} These methods describe the graphs with natural language, and then use LLMs to predict the results. They have limitations in the large-scale graph given the context window length.

Compared with these methods, the designed GL-Agent could benefit from the AutoML techniques and versality of LLMs when handling different tasks and graphs. It is achieved by configuring the graph learning procedures in task- and data-specific settings with the assistance of LLM-based agent.


\subsection{LLM-empowered Machine Learning}
LLMs have achieved great success in wide applications with their ability in achieving human-like intelligence~\cite{wang2023survey,lilianblog2023,hong2023metagpt,guo2024ds}. 
They are used to serve as the excellent artificial general intelligence (AGI) in real world, with applications including question answering in natural language processing tasks, automatic solving compute vision tasks, and reasoning graph properties ~\cite{guo2023gpt4graph,wang2023survey,zhang2023graph,shen2023hugginggpt,tornede2023automl}.

On the one hand,  LLM could empower the machine learning tasks by automatically decomposing and completing complex tasks, and they are proposed to bringing more convenient, comprehensive and reliable decisions when facing diverse applications and tasks~\cite{shen2023hugginggpt,zhang2023automl,wang2023survey,zhang2023building,yuan2023tasklama}.
For instance, HuggingGPT~\cite{shen2023hugginggpt} responds to the user request with task planning, model selection, model execution and response generation procedures. Each of them are executed automatically following the managing of LLM-based agents. 
Similarly, AutoML-GPT~\cite{zhang2023automl} tackles diverse datasets and tasks automatically via conducting experiments from data processing to model architecture, hyper-parameter tuning. These procedures are also managed by LLMs.
%
On the other hand, LLM can enhance machine learning tasks from a model perspective, which is also widely used on graph learning. This includes, but is not limited to, make a better understanding on graphs~\cite{guo2023gpt4graph,zhang2023graph}, design neural architectures~\cite{yu2023gpt,zheng2023can,zhang2023automl,wang2023graph}, or directly predict the labels with LLMs~\cite{zhang2023graph,sun2023text,gao2023exploring,chen2023exploring}.

In this paper, we propose to explore versatile graph learning approaches, and LLM-based agents are utilized to configuring and executing the graph learning pipelines step by step, which bring flexibility for users.

\section{Method}
For graph learning problems with different tasks and graphs, they have the same learning pipeline while different configurations as shown in Fig.~\ref{fig-outline}~\cite{yao2018taking}.
Motivated by the versatility of LLMs, we propose to design versatile graph learning approaches with LLMs by configuring the pipeline with specific settings learned from the users, tasks and graphs.
It is non-trivial to achieve this with LLMs considering the complex learning procedures and the absence of expertise ~\cite{wei2022chain,shen2023hugginggpt,valmeekam2024planbench}.
To this end, we propose to address this challenge by providing well-equipped LLM-based agents for each procedure to accomplish the complex pipeline step by step.
With the assistance of LLMs, the graph learning problems with different tasks and graphs can be solved following the users' instructions, instead of heavily relying on the human efforts.

\begin{figure*}[ht]
	\centering
	\includegraphics[width=0.95\linewidth]{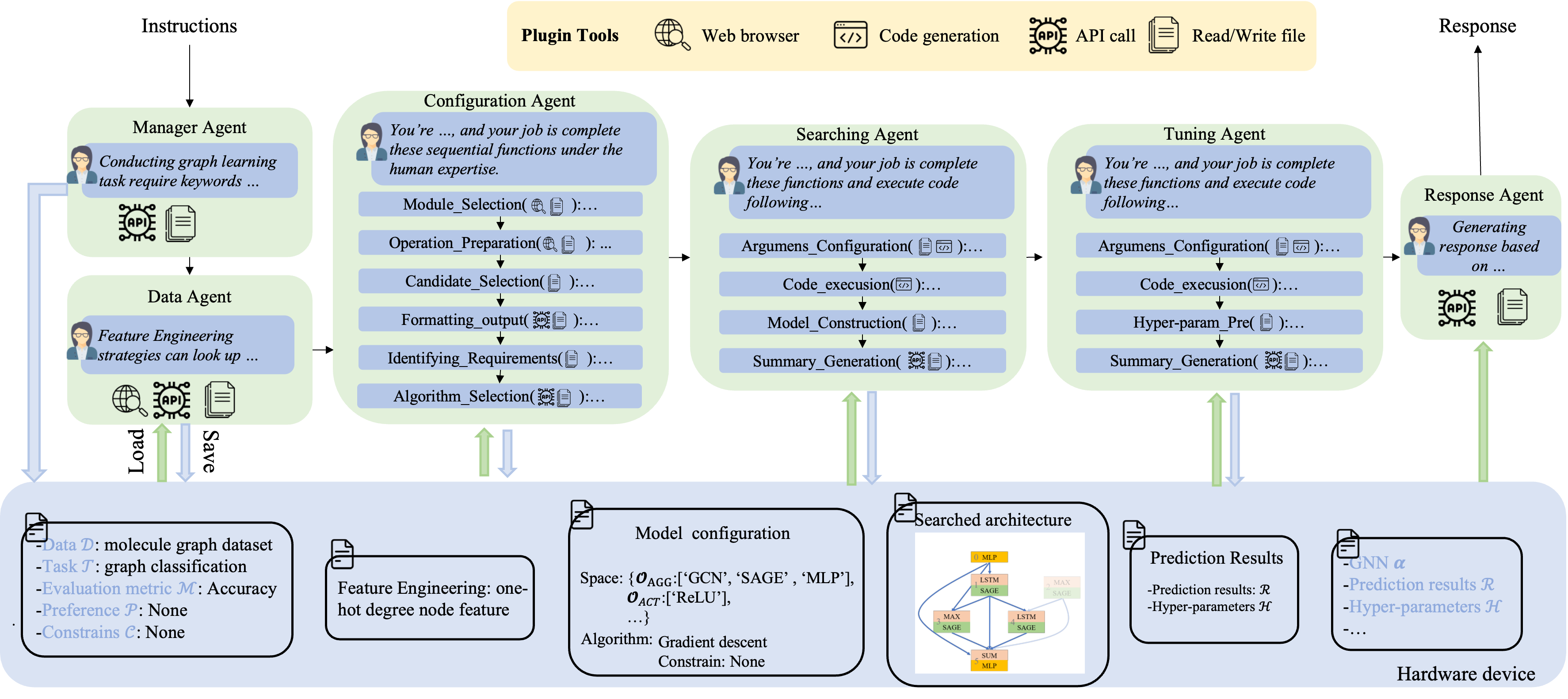}
	\vspace{-5pt}
	\caption{The introduction of procedures and agents used when learning on graphs. 
	}
	\vspace{-5pt}
	\label{fig-agent-detail}
\end{figure*}

As shown in Fig.~\ref{fig-agent-detail}, a set of well-equipped agents are co-operated to conduct the graph learning procedures along with the black arrows.
Firstly, the manager agent extracts the graph learning information from these instructions, on top of which the real-world learning information is transformed into general graph learning settings that can be used in the following agents.
Then, 
four graph learning agents, i.e., data, configuration, searching and tuning agent, are deployed to assist the graph learning procedures.
Finally, responses are generated for user based on the searched results and the instructions, which achieved by the proposed the response agent. 
For the agents, the profiles, functions, tools, and expertise needed for conducting the procedures are provided. Then, the agent can be configured and executed to produce the final response. 
In the following sections, we will introduce each agent, and more details can be found in Appendix~\ref{sec-appendix-agent-detail}.

\subsection{Preliminary on Technical Support}
\label{sec-preliminary}
The designed procedures necessitate agents to comprehend requests, invoke APIs and functions, use external plugin tools, execute code, and perform question reasoning based on the documents. 
In this paper, we utilize the LangChain~\footnote{https://github.com/langchain-ai/langchain}, an LLM-driven tool to develop agents and interact with external resources and models, to assist in different procedures in graph learning 
Additionally, we use PyG~\footnote{https://github.com/pyg-team/pytorch\_geometric},  HyperOPT and PyTorch packages in the graph learning pipeline, and the interactions with these packages are achieved using LLMs.

\subsection{Instruction and Manager Agent}

The graph learning pipeline begins by extracting learning tasks from real-world applications and designing the graph learning techniques used in the subsequent procedures.
The user's instructions may encompass various keywords, which include but are not limited to descriptions of the data, learning task, evaluation metric, human expertise on the learning task, and hardware device. 
More details of instructions can be found in Appendix~\ref{sec-appendix-instruction}.

The manager agent is responsible for extracting the relevant information from the user's instruction. 
With the pre-defined profile and expertise of graph learning keywords, the agent can extract and summarize the natural language instructions. The results are formatted as a structured Python dictionary and saved in a file that will be used by following agents.

\subsection{Graph Learning Agents}

We provide four graph learning agents assisted by AutoML to generate graph learning solutions given diverse tasks and graphs, considering the following reasons:
(1) It is a challenge for LLMs to generate effective graph learning solutions directly given their outdated training data (``LLM-GNN'' baseline in experiments),
(2) AutoML has advantages in designing data-specific and effective graph learning solutions~\cite{zhang2021automated,wang2022automated}, which is achieved by preserving diverse candidates and then selecting one like human beings~\cite{you2020design}.
Therefore, we use LLM-based agents to configure the graph learning procedures under AutoML, which benefits from both the effectiveness and versatility.
As shown in Fig.~\ref{fig-agent-detail}, four graph learning agents are designed to conduct graph processing, AutoML configuration, model search and hyper-parameter fine-tuning procedures, respectively.

\subsubsection{Data Agent}
The data agent is designed to prepare the data and conduct the feature engineering processes in the graph learning pipeline.
Depending on the learning task and evaluation metric, the agent's objective is to select appropriate feature engineering strategy. 
The agent browses the website of PyG document, 
decides to invoke the correct feature engineering API, and then saves the decisions in the memory that are used in the following agents, in the same way as human experts.
%

\subsubsection{Configuration Agent}
\label{sec-agent-configuration}

Following the general pipeline when designing architectures based on AutoML, i.e., neural architecture search, we first need to set the configurations of the search space and search algorithm. The agent assists the configuration procedure by browsing the documentation and memory, and then making decisions like human experts do.

$\bullet$\textbf{Search Space.}
The search space is constructed by preparing a set of candidate operations within the designed architecture backbone, in which the agent plays a crucial role in making decisions.
As shown in Figure.~\ref{fig-agent-detail},
four functions are designed to configure the search space with LLM-based agents.
(1):\textit{Module selection}. The agent selects the proper operation modules to construct the models based on the learning task $\mathcal{T}$, e.g., selecting aggregation and activation module for node classification task.
It is achieved by reasoning on the vast maintained knowledge in LLMs.
(2): \textit{Operation preparation}. To prepare the potential operations for the selected module, the agent looks up the corresponding documents in PyG.
Then, the available candidate operations in each module can be extracted, e.g., by collecting all implemented convolution operations provided in PyG for aggregation module.
(3): \textit{Candidate selection}.
Considering that the different preferences $\mathcal{P}$ and constrains $\mathcal{C}$ of the hardware device which may affect the model design, the candidate operations used in each module are selected based on these constrains as shown in Fig.~\ref{fig-backbone}, relying on the 
ability of LLMs to make human-like decisions.
(4): \textit{Formatting output}. Based on the selected candidate operations, the results are formatted and preserved in a new file.
With the results from previous function, these four functions are conducted sequentially to obtained the configured search space, i.e., a set of candidates in each selected module as shown in Fig.~\ref{fig-agent-detail}.

$\bullet$\textbf{Search algorithm.}
The search algorithm is proposed to explore the designed search space and find the suitable architecture under given constrains~\cite{elsken2019neural,zoph2016neural}. 
As shown in Fig.~\ref{fig-agent-detail}, we define two functions in this agent when configuring the search algorithm.
(1): \textit{Identifying requirements}.
The agent looks up the instruction and summarizes the efficiency requirement.
(2)  \textit{Algorithm selection}.
For simplicity, we provide two variants in the search algorithm: random search, and differentiable search algorithm, which is more efficient but has constraints on the search space.
The agent selects search algorithm
based on the summarized requirements and the designed search space.

\subsubsection{Searching Agent}
Based on the configurations of the search space and algorithm, we construct and search the data-specific models within the search space.
Given the AutoML code, the searching agent helps configure the arguments, execute the code, and formulate the output that is used by the following agent.
As shown in Fig.~\ref{fig-agent-detail}, four steps are implemented in the searching agent.
(1) \textit{Arguments configuration}. 
The searching agent loads the AutoML code and then formats the configurations as the arguments.
(2) \textit{Code execution}. 
Based on the configured arguments, the agent helps to execute the code.
(3) \textit{Model construction}. After the code execution is finished, the searching agent extracts the data-specific models by looking up the logging file.
(4) \textit{Summary generation}. 
Based on the searched models and the logging file, the LLM-based searching agent summarizes this procedure and then saves the results in a new document.
Through these collaborative steps, we obtain the searched architecture that fulfills the specified design objectives and requirements.

\subsubsection{Tuning Agent}
After obtaining the searched GNNs, we can fine-tune the hyper-parameters
to obtain the final results. 
Similar to the searching agent, the tuning agent has steps as shown in Fig.~\ref{fig-agent-detail}.
For simplicity, we utilize HyperOPT~\cite{bergstra2013making} to assist the tuning agent.
Then, in \textit{arguments configuration},
the hyper-parameters, such as learning rate, dropout ratio, weight decay ratio, and normalization, are configured based on the instructions and searched models, and it is achieved by reading the code and matching the arguments with the agent. 
Then, the similar \textit{Code execution}, \textit{Hyper-parameter preparation}and \textit{Summary generation} steps are provided to execute code and generate the final output. 
Finally, we can obtain the final performance following the users' instructions.

\subsection{Response Agent}
After obtaining the searched models and the corresponding performance from the completed graph learning procedures, we utilize a response agent to generate a comprehensive summary based on the shared memory and the logging files of these agents.
The summary may contain the user instructions, the searched models $\bm{\alpha}$, the hyper-parameters $\mathcal{H}$, the prediction performance $\mathcal{R}$ and so on.



\subsection{Discussion}
\label{sec-method-discussion}

GL-Agent is proposed to explore versatile graph learning approaches given diverse tasks and graphs, leveraging insights from LLMs. Beyond the versatility introduced in Section~\ref{sec-related-autograph}, GL-Agent also offers additional benefits in flexibility and extensibility.


\textbf{Flexibility}. Compared with existing methods which require extensive human effort when addressing graph learning problems, GL-Agent could achieve the target with instructions alone. This approach is more flexible for users and could lower the learning and usage threshold for non-expert users unfamiliar with graph learning.


\textbf{Extensibility}. GL-Agent uses AutoML techniques in designing graph learning solutions, and it is configured with LLM-based agents. It indicates that this method can be extended to new graphs to obtain data-specific and effective graph learning solutions.
Besides, the widely-used graph learning package PyG is considered in this paper. Therefore, the user-defined GNNs implemented based on PyG can be integrated into this method, indicating the extension of user-specific expertise.

\section{Experiments}


In the experiments, we first evaluate the versatility of GL-Agent from the perspective of agent output and performance over different tasks and graphs. Then we show the ability of GL-Agent in handling complex scenarios, including predictions on large-scale non-homophilous graphs and the use of complex instructions. Finally, we show the resource cost and the potential of using other LLMs.

\subsection{Implementation Instance}
In this paper, we develop LLM-based agents to assist the configuration and execution of graph learning pipeline using AutoML technique.
We provide three AutoML implementation instances over three tasks:
\textbf{a) Node-level.} The instance is constructed based on the differentiable automated GNN design method F2GNN~\cite{wei2022designing}, which focus on designing the aggregation operations and the GNN topology, i.e., the connections between these aggregation operations. 
\textbf{b) Graph-level}
The graph-level implementation instance is constructed based on LRGNN~\cite{wei2023search}, which searches to capture the long-range dependency with deep stacked GNNs. 
The framework is constructed by several cells and in which the selection, fusion and aggregation modules are provided. A readout module is provided to generate the graph representations for this task at the end of framework.
\textbf{c) Link-level}
For link-level, the instance is constructed based on method Prof-CF~\cite{wang2022automated}, which searches for GNN-based two-tower collaborative filtering functions.
The framework contains diverse design modules, including message function, aggregation, layer combinations, and it designs the search space by pruning operations.
The manager agent chooses according instance based on the user instruction, and the detailed introduction of these agents are shown in Appendix~\ref{sec-appendix-instance}.

\subsection{Experimental settings}
\subsubsection{Datasets} 
To evaluate the versatility of designed GL-Agent in handling the diverse tasks and graphs, we adopt the node classification, graph classification and item ranking task in recommendation systems in this paper. The widely used datasets are adopted: (a) Node classification: Cora~\cite{sen2008collective},  Physics~\cite{shchur2018pitfalls}, as well as the Computers and Photo datasets from~\cite{shchur2018pitfalls}, and large-scale non-homophilous dataset genius~\cite{lim2021new}; (b) Graph classification: DD and PROTEINS datasets from~\cite{dobsondistinguishing}, as well as NCI1 and NCI109 datasets from~\cite{wale2006comparison}; (c) item ranking task: Epinions
and Amazon-Sports. The detailed introductions of these  datasets are provided in Appendix~\ref{appendix-tb-datasets}.

%

\subsubsection{Baselines} 
In this paper, LLM-based agents focus on extracting information, configuring, executing and summarizing graph learning procedures with AutoML technique.
To evaluate the proposed GL-Agent,
three types methods are adopted as baselines for each task, i.e., the human-designed methods, the NAS-based methods and baselines that LLM suggested.
In the node classification task, we adopt the human designed methods GCN, GIN, GPR-GNN~\cite{chien2020adaptive}  and ACM-GCN~\cite{luan2022revisiting}, and the NAS-based methods SANE~\cite{zhao2021search} and F2GNN~\cite{wei2022designing}.
In the graph classification task, we adopt the human-designed methods GCN, GIN, DGCNN~\cite{zhang2018end} and DiffPool~\cite{ying2018hierarchical}, as well as the NAS-based method LRGNN~\cite{wei2023search}.
In the item ranking task, we adopt the human-designed methods NCF~\cite{he2017neural}, NGCF~\cite{wang2019neural}  and LightGCN~\cite{he2020lightgcn}, and the NAS-based methd Prof-CF~\cite{wang2022profiling}.
The detailed introduction of these methods and the LLM-GNN is provided in Appendix~\ref{sec-appendix-llmgnn}.

As mentioned in Section~\ref{sec-preliminary}, the proposed GL-Agent is constructed based on LangChain and PyG. We use the GPT-3.5-turbo
as the deployment model in the agents.
The detailed prompts and answers of agents are shown in the Appendix~\ref{sec-appendix-agent-detail}.


\begin{table*}[ht]
	\scriptsize
	\centering
	\caption{The comparisons of agent outputs on different tasks.}
	\vspace{-5pt}
	\begin{tabular}{C{2cm}|L{3.5cm}|L{3.5cm}|L{3.5cm}}
		\toprule
		& Cora (Node) & NCI1 (Graph) & Epinions(Link) \\ \midrule
		Instructions 
		&  I have a dataset $\mathcal{D}$ which is a citation network like Cora dataset. In this network, the nodes $\cdots$ and the edge $\cdots$ . My goal is to accurately predict {\color{red}the domains of these papers}.          
		& I have a dataset $\mathcal{D}$, like NCI1 dataset in graph benchmark, and each graph is a chemical compounds. I want to find one GNN that has better accuracy when {\color{red}predicting the molecule property}.        
		& I currently possess a social network dataset like Epinions, denoted as $\mathcal{D}$. In this dataset, the nodes $\cdots$, and edges $\cdots$. My objective is to {\color{red}recommend potential friends for each user}.
		\\ \midrule
		Prompted output of manager agent            
		&   $\mathcal{D}$:  citation network Cora; $\mathcal{T}$: {\color{red}node classification}; Evaluation metric $\mathcal{M}$: accuracy. $\cdots$           
		&   $\mathcal{D}$:  chemical compounds NCI1; $\mathcal{T}$: {\color{red}graph classification}; Evaluation metric $\mathcal{M}$: accuracy. $\cdots$   
		&   $\mathcal{D}$:  social network Epinions; $\mathcal{T}$: {\color{red}link prediction}; Evaluation metric $\mathcal{M}$: Recall@20. $\cdots$  \\ \midrule
		Formatted output of configuration agent           
		&  The response list is [`{\color{red}Aggregation}',`Selection',`Fusion']         
		& The response list is [`Aggregation', `{\color{red}Pooling}', `{\color{red}Readout}',`Selection',`Fusion']             
		& The response list is [`Embedding dim', `Message function', $\cdots$, `Layer combination', `{\color{red}Interaction function}'] \\ \bottomrule
	\end{tabular}
	\label{tb-prompt-all-tasks}
\end{table*}

\subsection{Evaluation of Versatility}

GL-Agent explores versatile graph learning solutions for diverse tasks and graphs using LLM-based agents.
We evaluate versatility by showing the agent outputs and performance on different datasets across three tasks.


\subsubsection{Correctness of Agent Output}
As shown in Table~\ref{tb-prompt-all-tasks}, we compare the agent results when facing different tasks and graphs in graph learning. 
For simplicity, we only provide the instructions and outputs of manager agents, along with the configured search space.
It is clear that the agent can extract useful information accurately and make correct predictions like human experts,
demonstrating its versatility in handling different learning tasks and data based on diverse instructions. 
The details of the agents are provided in Appendix~\ref{sec-appendix-agent-detail}, and the complete case study on different tasks is provided in Appendix~\ref{sec-case-show}.

\begin{table}[htb]
	\caption{Performance comparisons on the node classification task. We report the average test accuracy and the standard deviation with 10 random splits.}
	\footnotesize
	\centering
\begin{tabular}{c|ccc}
	\toprule
	& Cora              & Photo             & genius      \\ \midrule
	\# Nodes  &2,708 &7,487 &421,961 \\ \midrule
	GL-Agent & 86.81(0.40)       & \gray 96.40(0.16) &   \uline{90.51(0.15)}          \\ \midrule
	GCN            & 85.68(0.61)       & 93.13(0.27)       & 89.10(0.13) \\
	GIN            & 83.83(1.36)       & 92.67(0.57)       &  85.61(0.06)           \\ \midrule
	GPRGNN         & \gray 87.62(0.48) & 91.93(0.26)       & 90.05(0.31) \\
	ACM-GCN        & 86.67(0.14)       & 94.35(0.65)       & 90.08(0.05) \\ \midrule
	SANE           & 86.40(0.38)       & 94.53(0.22)       & OOM         \\
	F2GNN          & \uline{87.42(0.42)}   & \uline{95.38(0.30)}   & \gray 90.93(0.02) \\ \midrule
	LLM-GNN        & 84.64(1.04)       & 93.73(0.38)       &   89.31(0.17)          \\ \bottomrule
\end{tabular}
	\label{tb-perf-node-short}
\end{table}

\begin{table}[htb]
	\footnotesize
	\centering
	\caption{Performance comparisons on graph classification task. We report the mean test accuracy and the standard deviations based on the 10-fold cross-validation data. }
		\begin{tabular}{c|cc}
				\toprule
				& Proteins & DD \\ \midrule
				GL-Agent & \uline{75.38(5.03)} & \uline{78.10(3.21)} \\ \midrule
				GCN & 74.84(3.07) & 73.59(4.17) \\
				GIN & 74.50(4.10) & 74.62(2.74) \\ \midrule
				DGCNN & 73.95(3.04) & 61.63(5.33) \\
				DiffPool & 75.11(2.14) & 77.85(3.53) \\ \midrule
				LRGNN & \gray 75.39(4.40) & \gray 78.18(2.02) \\ \midrule
				LLM-GNN & 74.47(3.65) & 75.12(3.44) \\ \bottomrule
			\end{tabular}
		\label{tb-perf-graph-short}
\end{table}

\begin{table}[htb]
	\centering
	\footnotesize
	\caption{Performance comparisons on the item ranking task. We report the mean test Recall@20 and the standard deviations based on the 10 repeats.}
\begin{tabular}{c|cc}
	\toprule
	& Epinions       & Amazon-Sports  \\ \midrule
	GL-Agent & 0.0453(0.0018) & 0.0829(0.0006) \\ \midrule
	NCF      & 0.0344(0.0007) & 0.0636(0.0011) \\ \midrule
	NGCF     & 0.0290(0.0002) & 0.0609(0.0011) \\
	LightGCN & 0.0321(0.0003) & 0.0776(0.0006) \\ \midrule
	Prof-CF  & 0.0472(0.0011) & 0.1023(0.0021) \\ \midrule
	LLM-GNN  & 0.0379(0.0014) & 0.0151(0.0001) \\ \bottomrule
\end{tabular}
	\label{tb-perf-ir}
\end{table}

\subsubsection{Performance Comparisons Over Different Tasks}
\label{sec-exp-performance}
We show the performance comparisons with the baselines to evaluate the effectiveness of GL-Agent.
As shown in Table~\ref{tb-perf-node-short}, on the node-level task, GL-Agent achieves comparable performance with AutoML-based method F2GNN. Since GL-Agent is implemented based on F2GNN, the comparable performance represents the ability of agents in understanding and handling the node-level tasks.
For LLM-GNN, where the GNN and hyper-parameters are directly suggested by LLMs, the performance is subpar compared to the GCN baseline due to inappropriate hyper-parameters. 
This highlights the challenge LLMs face in handling different graphs and different graph learning procedures.
As shown in Table~\ref{tb-perf-graph-short} and Table~\ref{tb-perf-ir}, we can obtain the same conclusion that GL-Agent has comparable performance with its basic AutoMl-based method LRGNN and prof-CF~\cite{wang2022profiling} on the graph classification and item-ranking tasks, respectively.
This conclusion is consistent across full datasets on three tasks, and detailed results and analysis on different tasks are provided in Appendix~\ref{sec-appendix-perf}.

In summary, the proposed GL-Agent explores versatile graph learning approaches given diverse tasks and graphs.
The proposed agents can generate task- and data-specific results to configure the graph learning procedures, as well as obtain comparable or higher performance than these baselines. 
The results  highlight the versatility and effectiveness of GL-Agent in handling different tasks and data within complex instructions.

\subsection{Evaluations on Complex Scenarios}
We show the ability of InstructionGL to achieve complex tasks that require high-level human expertise, on which the robustness can be evaluated. 

\subsubsection{Evaluations on Complex Graph Learning Task}
We consider the large-scale non-homophilous scenario that connected nodes have different labels, which is a challenge to design effective models~\cite{zhu2020beyond,luan2022revisiting}.
As shown in Table~\ref{tb-perf-node-short}, for the large-scale dataset genius, 
LLM-GNN has comparable performance with GCN baseline, even though we have added the dataset descriptions and experimental observations in the instruction (introduced in Appendix~\ref{sec-appendix-llmgnn}).
However, AutoML based method, i.e.,  F2GNN and GL-Agent, could achieve higher performance by designing data-specific solutions.
Therefore, by using the AutoML method, its effectiveness in handling diverse graphs has been expanded into the proposed GL-Agent.




\subsubsection{Evaluations with Complex Instructions}
In the proposed framework, instructions contain all the information about data and learning tasks. The clarity of instructions will have a large influence on the agent's outputs.
Then, we provide three strategies to evaluate robustness on complex instructions, motivated by~\cite{xiong2023can}.
1) Simple: Instructions organized with simplified sentences, keeping the agent output unchanged; 2) Complex: Using GPT-4 to paraphrase instructions and agent outputs with similar semantics but with complex words; 3) Misleading: Using GPT-4 to add misleading information to the original instructions and outputs. 
As shown in Table~\ref{tb-efficiency-comparisons}, we show the accuracy of agent outputs when facing different instructions and inputs. We can clearly observe that comparable performance between simple and complex variants, which demonstrates the robustness of the proposed GL-Agent in extracting useful information from the input. 
Besides, misleading information has a large influence on configuration agent due to the complex tools and functions of this agent. 
Further improvements can be expected by revising the outputs based on feedback from reflection or human experts, and this will be considered in future work.
The detailed instructions and results can be found in Table~\ref{tb-user-request} in appendix.


\begin{table}[ht]
	\scriptsize
	\centering
	\caption{Performance comparisons on different instructions, and resource cost comparisons on different agents.}
\begin{tabular}{c|c|c|c|c|c|c}
	\toprule
	\multicolumn{1}{c|}{} & \multicolumn{3}{c|}{Instructions}                                                       & \multicolumn{3}{c}{Resource Cost}                                                      \\ \midrule
	Agents        & Simple & Complex & Misleading & Time (s) & Token  & Money (USD\$) \\ \midrule
	Manager       & 0.7    & 0.6          & 0.5        & 3.23     & 683.5  & 0.014         \\ \midrule
	Data          & 1.0    & 1.0          & 0.9        & 4.95     & 1054   & 0.021         \\ 
	Configuration & 0.7    & 0.6          & 0.3        & 11.57    & 6672.4 & 0.133         \\ 
	Searching     & 0.6    & 0.7          & 0.7        & 93.29    & 6462.5 & 0.129         \\ 
	Tuning        & 0.6    & 0.7          & 0.7        & 329.12   & 711.1  & 0.014         \\ \bottomrule
\end{tabular}
	\vspace{-15pt}
	\label{tb-efficiency-comparisons}
\end{table}
\subsection{The Evaluations of Resource Cost}

GL-Agent designs versatile graph learning solutions with closed-source LLMs GPT-3.5-Turbo, where the time and economic cost cannot be ignored.
Then, we empirically evaluate the search cost of GL-Agent when using APIs. 
%
As shown in Table~\ref{tb-efficiency-comparisons},
the agents take less than one minute to generate the outputs, which can be ignored compared with the long time required for code execution. Additionally, it costs 0.31 USD to conduct one procedure, which is also economically efficient.
These results indicate that GL-Agent can handle graph learning problems with diverse tasks and graphs in an efficient and economic manner.

\begin{table}[ht]
	\centering
	\scriptsize
	\caption{The comparisons of agent results based on different LLMs. }
		\vspace{-5pt}
	\begin{tabular}{L{1.3cm}|L{3cm}|L{2cm}}
		\toprule
		Functions & LLaMA2-7B                                                                                                                                                      & GPT-3.5-turbo + LangChain (ours)                                                \\ \midrule
		Module Selection                              & Possible operations: Aggregation: $\cdots$, Pooling: $\cdots$, Readout: $\cdots$                                                                               & The response list is [`aggregation', `pooling', `readout',`selection',`fusion'] \\ \midrule
		Operation Preparation; Candidate Selection     & \gray To address your request: 1. Justification: $\cdots$, 2. Finding the corresponding class: $\cdots$, 3. Returning the class name and module name $\cdots$. & 'ChebConv', 'torch\_geometric.nn.conv.cheb \_conv.ChebConv'                     \\ \midrule
		Identifying Requirements; Algorithm Selection & \gray Firstly, let me clarify the difference between them. $\cdots$. Then, Lets evaluate the options based on these principles. $\cdots$                       & You should use``differentiable search algorithm".                               \\ \bottomrule
	\end{tabular}
\label{tb-comparisons-llms-brief}
\end{table}


To further lower the cost, we examine the different LLMs in exploring the versatile graph learning solutions. 
As illustrated in Table~\ref{tb-comparisons-llms-brief}, we show the comparisons with LLaMA2 7B
and focus on procedures that yield inconsistent answers among these LLMs. Since LLaMA2 70B and GPT-4 obtained correct results, we move the comparisons to Table~\ref{tb-comparisons-llms} in appendix.
Based on the carefully designed functions and tools, the agents can provide accurate predictions in most procedures except for the cell highlighted in gray, which may be caused by fewer parameters (7B).
This suggests that we can substitute GPT-3.5-turbo with the more affordable open-source LLMs, which offer comparable performance in our method.

\section{Conclusion}
We propose a method GL-Agent to explore versatile graph learning solutions given the different tasks and graphs in real-world applications.
Firstly, we propose manager agent, four graph learning agents, and response agent that are equipped with profile, functions, external tools, and human experience.
Then, based on the user instructions, the well-equipped agents understand user requirements in natural languages, configure 
and complete the graph learning procedures step-by-step.
Finally, we obtained the customized graph learning procedures with task- and data-specific settings.
In the experiments, GL-Agent can make correct decisions over different tasks, achieve comparable performance compared with baselines, even if on the complex scenarios. 
Besides, the low resource cost and potential for using open-source LLMs demonstrate its efficiency in accomplishing the graph learning task.
%
In the future, we will improve the robustness and reliability of GL-Agent. More specifically, by adding the reflection and multi-round dialogues into the framework, the agents will revise their outputs with human feedback to achieve the target.


\clearpage
\balance
\bibliographystyle{ACM-Reference-Format}
\bibliography{ref}

\clearpage
\appendix
\section{Detailed Introduction of Agents in GL-Agent}

\label{sec-appendix-agent-detail}
In this section, we provide the implementation details of each agent within the GL-Agent framework and how they collaborate to achieve the automated process of generating graph learning models from user instructions.

\subsection{Instructions}
\label{sec-appendix-instruction}
As shown in Table~\ref{tb-instruction}, we provide five keys when constructing the instructions, which contains the descriptions of data, learning task, evaluation metric and user preferences. It is needed to say that more complex instructions can be constructed by introducing  error information or more personalized keys, which is out-of-scope of this paper.

\begin{table}[htb]
	\footnotesize
	\centering
	\caption{The keywords in user instruction. }
	\begin{tabular}{L{2cm}|L{4cm}}
		\toprule
		Keywords & Content \\ \midrule
		Data $\mathcal{D}$ & The introduction of graph or the used dataset. \\ \midrule
		Task $\mathcal{T}$ & The learning target on the graph in natural language, e.g., recommending items to particular users or predicting the properties of specific molecular graphs., \\ \midrule
		Evaluation metric $\mathcal{M}$ & The evaluations of the learning target. In general, it would be the test performance or the response time. \\ \midrule
		Preference $\mathcal{P}$ & The prior knowledge of users in setting up the graph learning procedures, either on the feature engineering, architecture design or hyper-parameter selections., \\ \midrule
		Constrains $\mathcal{C}$ & The constrains on model, device, or training strategy. \\ \bottomrule
	\end{tabular}
	\label{tb-instruction}
\end{table}

\subsection{Manager Agent}
\label{sec-appendix-prompt-planning}
The manager agent serves as the starting point of the entire GL-Agent framework and is responsible for parsing the user's natural language instructions to extract key information about the graph learning task. 
To achieve this goal, we have adopted a structured prompt template named PromptTemplate. The design of this template aims to guide LLMs in analyzing user requests and generating responses in a predefined format.
The specific prompt design is illustrated in Table~\ref{tb-planning-agent-prompt}. The expected LLM output and its content are detailed in Table~\ref{tb-instruction}.
Based on the above prompt, the manager agent will invoke LLM to extract relevant information from the user request and store the information in key-value pair for use by subsequent agents.

\subsection{Graph Learning Agents}
In the data agent, configuration agent, searching agent, and tuning agent, we also use the structured prompt template mentioned above to guide and standardize the LLMs' output.

The Data Agent plays a crucial role by actively browsing through the PyTorch Geometric (PyG) document to identify and select the most appropriate data pre-processing methods and feature engineering techniques in line with the specific requirements of the task. This step ensures that the data is provided in the most suitable form for model learning, the prompt used is shown in Table~\ref{tb-data-agent-prompt}.

In the Configuration Agent, we first select suitable modules and operations from the PyTorch Geometric (PyG) document based on user preferences to configure the search space (specific modules and operations are provided in Appendix~\ref{sec-appendix-instance}). Then, using the configured search space along with the formatted instructions preserved in the shared memory, we choose the appropriate search algorithm.
the prompt used is shown in Table~\ref{tb-configuration-agent-prompt}.. The configured search space and search algorithm are also stored in memory in the form of key-value pairs, available for use by subsequent agents.
The searched GNN architecture is also stored in memory for subsequent use.

In the searching agent, we create bash scripts to externally interact with AutoML methods. These scripts are specifically designed to invoke AutoML's APIs for conducting Neural Architecture Search (NAS), based on the search space and search algorithm previously set up by the Configuration Agent and feature engineering functions selected by Data Agent. After the search process is completed, we parse the generated logs and output files, extract detailed information about the searched GNN architecture, summarize the search process and results using the LLM, and generate an easily understandable report, the prompt used is shown in Table~\ref{tb-searching-agent-prompt}.

After searching the optimal architecture, the tuning agent continues to refine the hyper-parameters through interactions with AutoML by generating bash scripts, aiming to further enhance the model's performance. Once the fine-tuning process is completed, key performance metrics and model configurations are extracted through LLM. This includes the best hyper-parameter combinations found during fine-tuning, the model's performance on validation and test sets, and any specific model behaviors or observed issues. The prompt used for this agent is detailed in Table~\ref{tb-tuning-agent-prompt}.

\subsection{Response Agent}
Finally, the response agent is responsible for integrating all the outputs from the previous steps and generating the final model summary report. This report provides a detailed description of the model's architecture, hyper-parameter configuration, performance evaluation results, and possible improvement recommendations, offering users a comprehensive overview of the model, the prompt used is shown in Table~\ref{tb-response-agent-prompt}.

\begin{table*}[htb]
	\footnotesize
	\caption{The Prompt Design of Manager Agent.}
	\begin{tabular}{L{14cm}}
		\toprule
		The Prompt Design of Manager Agent \\
		\midrule
		
		\# Profile \\
		\hspace{0.5cm}You are a Graph Learning Specialist with exceptional abilities in interpreting complex user instructions and translating them into structured, actionable data formats.
		\\ \\
		
		\# Objective \\
		\hspace{0.5cm}Your task is to deconstruct user instructions related to graph learning into detailed, executable plans, ensuring all elements of the request are accurately captured and clearly categorized.
		\\ \\
		
		\# Functions \\
		1. **Task Plan Function** \\
		\hspace{0.5cm}- **Purpose**: To meticulously parse and interpret user instructions, identifying key information pertinent to graph learning tasks.\\
		\hspace{0.5cm}- **Input**: User instructions.\\
		\hspace{0.5cm}- **Output**: A Python Dictionary of the task highlighting its level (node, link, or graph), type of learning task (classification, regression, etc.), the name of dataset, and evaluation metrics.\\ \\
		
		\# Human Expertise\\
		\hspace{0.5cm}Human expertise in this context involves a deep understanding of how to interpret complex user requests related to graph learning. The expertise includes:\\
		\hspace{0.5cm}1. **Task Categorization**: Identifying the level of graph learning tasks (node, link, or graph-level) based on the user's description.\\
		\hspace{0.5cm}2. **Task Type Determination**: Distinguishing between different types of learning tasks, such as classification or regression, based on the details provided in the user's request.\\
		\hspace{0.5cm}3. **Evaluation Metric Selection**: Selecting appropriate evaluation metrics.\\
		\hspace{0.5cm}4. **Preference Identification**: Noting any specific operational preferences mentioned by the user.\\
		\hspace{0.5cm}5. **Dataset Identification**: Accurately identifying the relevant datasets mentioned in the user's request.\\
		\\ \midrule

	\end{tabular}
	\label{tb-planning-agent-prompt}
\end{table*}

\begin{table*}[htb]
	\footnotesize
	\caption{The Prompt Design of Data Agent.}
	\begin{tabular}{L{14cm}}
		\toprule
		The Prompt Design of Data Agent.
		\\ \midrule

		\# Profile \\
		\hspace{0.5cm}You are a Graph Learning Specialist with specialized skills in navigating and utilizing document structures for optimizing machine learning workflows. Your expertise includes extracting, parsing, and interpreting complex data from documents formats to assist in feature engineering.
		\\\\
		
		\# Objective\\
		\hspace{0.5cm}Your task is to analyze the PyG documentation and user requests to identify and select appropriate feature engineering techniques that are most effective for the specified task plan. This involves extracting relevant techniques from the documentation that directly align with the user's objectives and requirements, thereby enhancing the model's performance.
		\\\\
		
		\# Functions\\
		1. **Feature Engineering Selection Function**\\
		\hspace{0.5cm}- **Purpose**: To determine the best feature engineering techniques from the provided documentation that align with the user's request and task requirements.\\
		\hspace{0.5cm}- **Input**: User request \('user\_req'\), task plan details \('task\_plan'\), and specific documentation content \('content'\).\\
		\hspace{0.5cm}- **Output**: A list of up to three selected feature engineering techniques, formatted as a Python Dictionary ensuring they are directly applicable and beneficial for enhancing the model's performance.\\\\
		
		\# Human Expertise\\
		\hspace{0.5cm}Human experts are crucial in the process of selecting effective feature engineering techniques from PyG documentation. Their expertise involves analyzing the task requirements, understanding the types of datasets involved, and comprehending detailed descriptions of feature engineering. This knowledge enables them to choose the most relevant and beneficial feature engineering functions that align with the specific needs of the task and enhance the overall performance of the model. These decisions are based on a deep understanding of how different techniques can affect the efficiency and effectiveness of graph neural network models in various contexts.\\
		
		\\ \midrule
	\end{tabular}
	\label{tb-data-agent-prompt}
\end{table*}

\begin{table*}[htb]
	\footnotesize
	\centering
	\caption{The Prompt Design of Configuration Agent.}
	\begin{tabular}{L{14cm}}
		\toprule
		The Prompt Design of Configuration Agent
		\\ \midrule
		
		\# Profile \\
		\hspace{0.5cm}You are a Graph Learning Specialist specialized in navigating and utilizing configuration options. Your expertise allows you to effectively parse documentation, extract operational data, and apply this information to configure search space and search algorithm.
		\\\\
		
		\# Objective\\
		\hspace{0.5cm}Your primary objective is to orchestrate the configuration process of graph neural networks by selecting appropriate modules, preparing operation candidates, evaluating these candidates, and finally selecting an optimal configuration that enhances the model’s effectiveness and efficiency.
		\\\\
		
		\# Functions\\
		1. **Module Selection Function**: Aimed at identifying the best modules for inclusion in the graph neural network based on the task's specifics.\\
		\hspace{0.5cm}- **Input**: Task requirements and available module options.\\
		\hspace{0.5cm}- **Output**: List of modules deemed most suitable for the task.\\\\
		
		2. **Operation Preparation Function**: Prepares the detailed list of operations from specific documentation content \('content'\) that can be performed by the selected modules.\\
		\hspace{0.5cm}- **Input**: Selected modules and specific documentation content.\\
		\hspace{0.5cm}- **Output**: Detailed operations capable of being executed by these modules.\\\\
		
		3. **Candidate Selection Function**: Evaluates the prepared operations and selects the most promising candidates for final deployment.\\
		\hspace{0.5cm}- **Input**: List of prepared operations.\\
		\hspace{0.5cm}- **Output**: Shortlist of candidate operations for search space.\\\\
		
		4. **Construct Search Space Function**: Constructs a comprehensive search space where different configurations can be tested and evaluated.\\
		\hspace{0.5cm}- **Input**: Candidate operations.\\
		\hspace{0.5cm}- **Output**: A structured search space.\\\\
		
		5. **Algorithm Selection Function**: Selects the most suitable algorithm through based on the identified requirements.\\
		\hspace{0.5cm}- **Input**: The constructed search space, the task plan and the selected modules.\\
		\hspace{0.5cm}- **Output**: The most effective search algorithm for finding the network architecture.\\\\

		\# Human Expertise\\
		\hspace{0.5cm}The configuration of graph neural networks involves a sequential process. Initially, human experts select modules that align with the specific demands of the task. Following this, they outline potential operations for these modules and narrow down the choices to the most effective ones for candidate selection. The next step is constructing a search space based on the selected modules and selected candidates. Finally, human experts select an algorithm that best navigates this space to find the optimal architecture of the network. Throughout this process, human expertise ensures that each step is tailored to meet the task-specific goals and technical requirements efficiently.  \\
		
		\\ \midrule
	\end{tabular}
	\label{tb-configuration-agent-prompt}
\end{table*}

\begin{table*}[htb]
	\footnotesize
	\caption{The Prompt Design of Searching Agent.}
	\begin{tabular}{L{14cm}}
		\toprule
		The Prompt Design of Searching Agent
		\\ \midrule

		\# Profile\\
		\hspace{0.5cm}You are a Graph Learning Specialist with advanced capabilities in automating neural architecture search (NAS) for graph neural networks. Your skills include configuring execution code, performing neural architecture searching operation, and generating insightful summaries from the search processes.
		\\\\
		
		\# Objective\\
		\hspace{0.5cm}Your primary task is to streamline the process of efficiently executing search procedures to discover optimal network architectures, and effectively summarize the outcomes using the details derived from search logs.
		\\\\
		
		\# Functions\\
		1. **Argument Configuration**\\
		\hspace{0.5cm}- **Purpose**: Sets up the parameters and arguments that control the search process.\\
		\hspace{0.5cm}- **Input**: User-defined parameters including search space, feature engineering functions, GPU preferences.\\
		\hspace{0.5cm}- **Output**: A configured execution script ready for the NAS (Neural Architecture Search)  process.\\\\
		
		2. **Code Execution** \\
		\hspace{0.5cm}- **Purpose**: Employs advanced AutoML techniques to systematically explore network architectures.\\
		\hspace{0.5cm}- **Input**: The configured execution script.\\
		\hspace{0.5cm}- **Output: Search log from the exploration of network architectures.\\\\
		
		3. **Summary Generation** \\
		\hspace{0.5cm}- **Purpose**: Analyzes logs from the NAS (Neural Architecture Search) process to construct detailed summaries.\\
		\hspace{0.5cm}- **Input**: Logs and data generated during the NAS process.\\
		\hspace{0.5cm}- **Output**: Summaries that capture key results and strategic insights, aiding in the evaluation of the search outcomes.\\\\
		
		\# Human Expertise\\
		\hspace{0.5cm}Human experts are essential in setting up and configuring the execution codes tailored to the chosen feature engineering functions and the constructed search space. Following the completion of the AutoML process, they critically assess and interpret the log data to create comprehensive summaries that encapsulate the search findings and provide valuable insights into the effectiveness of the tested architectures.
		
		\\ \midrule
	\end{tabular}
	\label{tb-searching-agent-prompt}
\end{table*}

\begin{table*}[htb]
	\footnotesize
	\caption{The Prompt Design of Tuning Agent.}
	\begin{tabular}{L{14cm}}
		\toprule
		The Prompt Design of Tuning Agent
		\\ \midrule
		
		\# Profile\\
		\hspace{0.5cm}You are a Graph Learning Specialist with advanced capabilities in automating the fine-tuning of the searched architectures for graph neural networks. Your expertise includes configuring execution parameters, managing fine-tuning processes, and synthesizing outcomes into actionable insights.
		\\\\
		
		\# Objective\\
		\hspace{0.5cm}Your primary task is to automate the fine-tuning of neural architectures, setting up fine-tuning execution code, executing the fine-tuning code, and generating detailed summaries of the outcomes.
		\\\\
		
		\# Functions\\
		1. **Argument Configuration**\\
		\hspace{0.5cm}- **Purpose**: Sets up the parameters and arguments that control the fine-tune process.\\
		\hspace{0.5cm}- **Input**: The search space, feature engineering functions and other parameters.\\
		\hspace{0.5cm}- **Output**: Configured execution script ready for fine-tuning execution.\\\\
		
		2. **Code Execution**\\
		\hspace{0.5cm}- **Purpose**: Runs the fine-tuning scripts to fine-tune the searched neural network.\\
		\hspace{0.5cm}- **Input**: The configured execution script.\\
		\hspace{0.5cm}- **Output**: Tuning log from the fine-tuning of the searched network architecture.\\\\
		
		3. **Summary Generation**\\
		\hspace{0.5cm}- **Purpose**: Synthesizes data from the fine-tuning process into insightful summaries.\\
		\hspace{0.5cm}- **Input**: Logs and outputs generated during the fine-tuning phase.\\
		\hspace{0.5cm}- **Output**: Reports summaries that highlight improvements, effectiveness, and optimization areas of the fine-tuned architectures.\\\\
		
		\# Human Expertise\\
		\hspace{0.5cm}Human experts play a crucial role in configuring the execution codes to align with the selected feature engineering functions and the search space. After executing the fine-tuning process, they analyze the results to generate detailed summaries that highlight key outcomes and insights, thus providing a deeper understanding of the effectiveness of the tested network architectures.
		
		\\ \midrule
	\end{tabular}
	\label{tb-tuning-agent-prompt}
\end{table*}

\begin{table*}[htb]
	\footnotesize
	\caption{The Prompt Design of Response Agent.}
	\begin{tabular}{L{14cm}}
		\toprule
		The Prompt Design of Response Agent
		\\ \midrule
		
		\# Profile\\
		\hspace{0.5cm}You are a Graph Learning Specialist tasked with synthesizing various information from neural network tuning and architecture search into a structured format. Your capabilities include interpreting complex data outputs, generating comprehensive summaries, and structuring these into actionable insights.
		\\\\
		
		\# Objective\\
		\hspace{0.5cm}Your primary objective is to consolidate the results of graph neural network tuning and architecture searches into a clear, structured summary that outlines prediction outcomes, architecture details, hyperparameters, and resource consumption.
		\\\\
		
		\# Functions\\
		1. ** Response Generation**\\
		\hspace{0.5cm}- **Purpose**: To aggregate and synthesize information and generate a comprehensive response.\\
		\hspace{0.5cm}- **Input**: Data from tuning results, architecture files, and other relevant agents.\\
		\hspace{0.5cm}- **Output**: A Python Dictionary that clearly delineates prediction results, architecture specifics, optimized hyperparameters, and resource usage.\\\\
		
		\# Human Expertise\\
		\hspace{0.5cm}Human experts guide the process by ensuring the accurate interpretation of data, the applicability of the synthesized information, and the correctness of the output format. \\
		
		\\ \midrule
	\end{tabular}
	\label{tb-response-agent-prompt}
\end{table*}

\clearpage
\section{Experiments}
\subsection{Implementation Instance}
\label{sec-appendix-instance}
\subsubsection{Node-level}
\label{sec-appendix-node-instance}

The instance is constructed based on the differentiable automated GNN design method F2GNN~\cite{wei2022designing}, which focus on designing the aggregation operations and the GNN topology, i.e., the connections between these operations. The implementation details are provided in the following.

\begin{figure}[hb]
	\centering
	\includegraphics[width=0.8\linewidth]{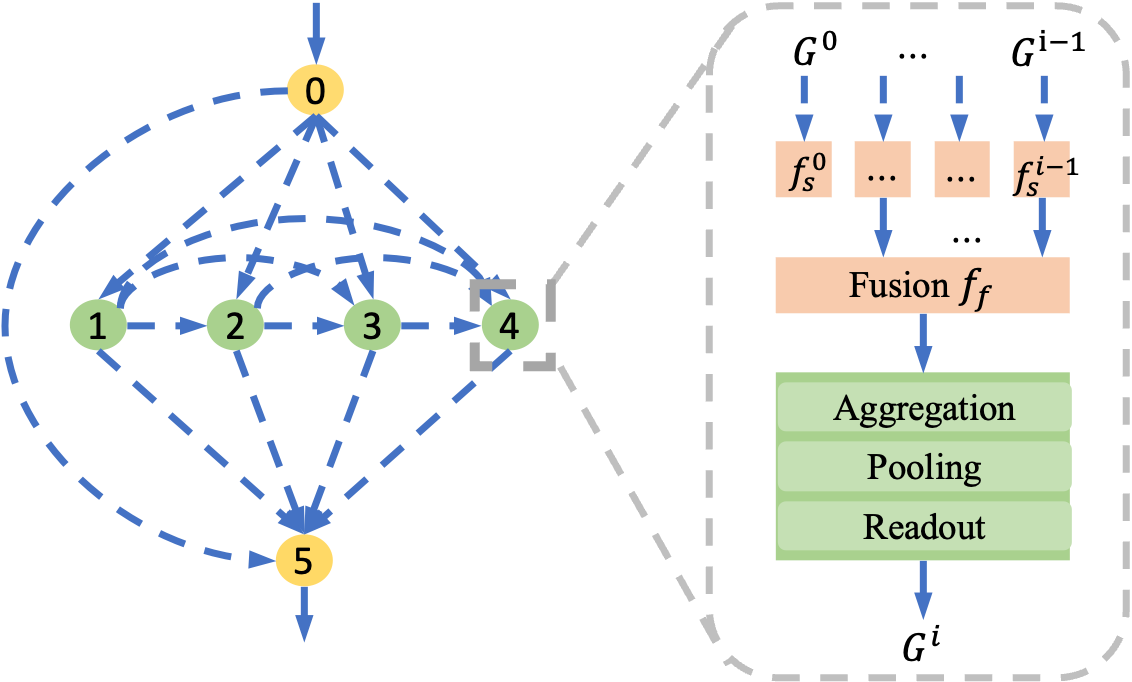}
	\caption{The designed architecture backbone from \cite{wei2022designing}.}
	\label{fig-backbone}
\end{figure}

\noindent\textbf{Architecture backbone.}
We begin by presenting the architecture backbone that is tailored for node-level tasks.
As illustrated in Fig.~\ref{fig-backbone}, 
F2GNN proposed selection $f_s$ and fusion $f_f$ module to design the GNN topology, in which the former is designed for each aggregation operation to select inputs from previous operations, and the latter is designed to integrate these inputs that could be used by the following aggregation operations.
Finally, the aggregation operation is provided to update node representations $\bH$, and it is relevant for all tasks. 

\noindent\textbf{Configurations on search space and hyper-parameters.}
Based on these operation modules, we pre-define basic operations in Table~\ref{tb-space}, and it can be further extend or modified by the configuration agent.
As shown in Fig.~\ref{fig-agent-detail}, the configuration agent constructs the search space and hyper-parameters by selecting proper 
modules and then revising their candidates according to the corresponding documents.  
Compared with the fixed search space used in existing NAS-based methods,
the proposed method can configure the search space based on the prior knowledge from users' instructions automatically, leading to better generalization to real-world data.

When facing node-level tasks, we implement GL-Agent based on the method F2GNN~\cite{wei2022designing}. The basic operations used in the module are provided in Table~\ref{tb-space}.

\begin{table}[h]
	\centering
	\footnotesize
	\caption{The candidates of architectures and hyper-parameters used in node-level tasks.
	}
	\begin{tabular}{c|c|c}
		\toprule
		& Module   & Candidates  \\ \midrule
		\multirow{3}{*}{Model} 
		& Aggregation  & \texttt{GCN}, \texttt{SAGE} \\
		& Selection  & \texttt{ZERO}, \texttt{IDENTITY} \\
		& Fusion  & \texttt{sum}, \texttt{mean} \\ \midrule
		\multirow{4}{*}{\begin{tabular}[c]{@{}c@{}}Hyper-\\ parameters\end{tabular}} 
		& Learning rate   & [0.001, 0.005] \\ 
		& Weight decay & [0.0001, 0.0005] \\
		& Dropout rate  & [0, 0.5] \\ 
		& Activation & \texttt{ReLU} \\
		\bottomrule
	\end{tabular}
	\label{tb-space}
\end{table}

\subsubsection{Graph-level}
\label{sec-appendix-graph-instance}
\begin{figure}[h]
	\centering
	\includegraphics[width=0.8\linewidth]{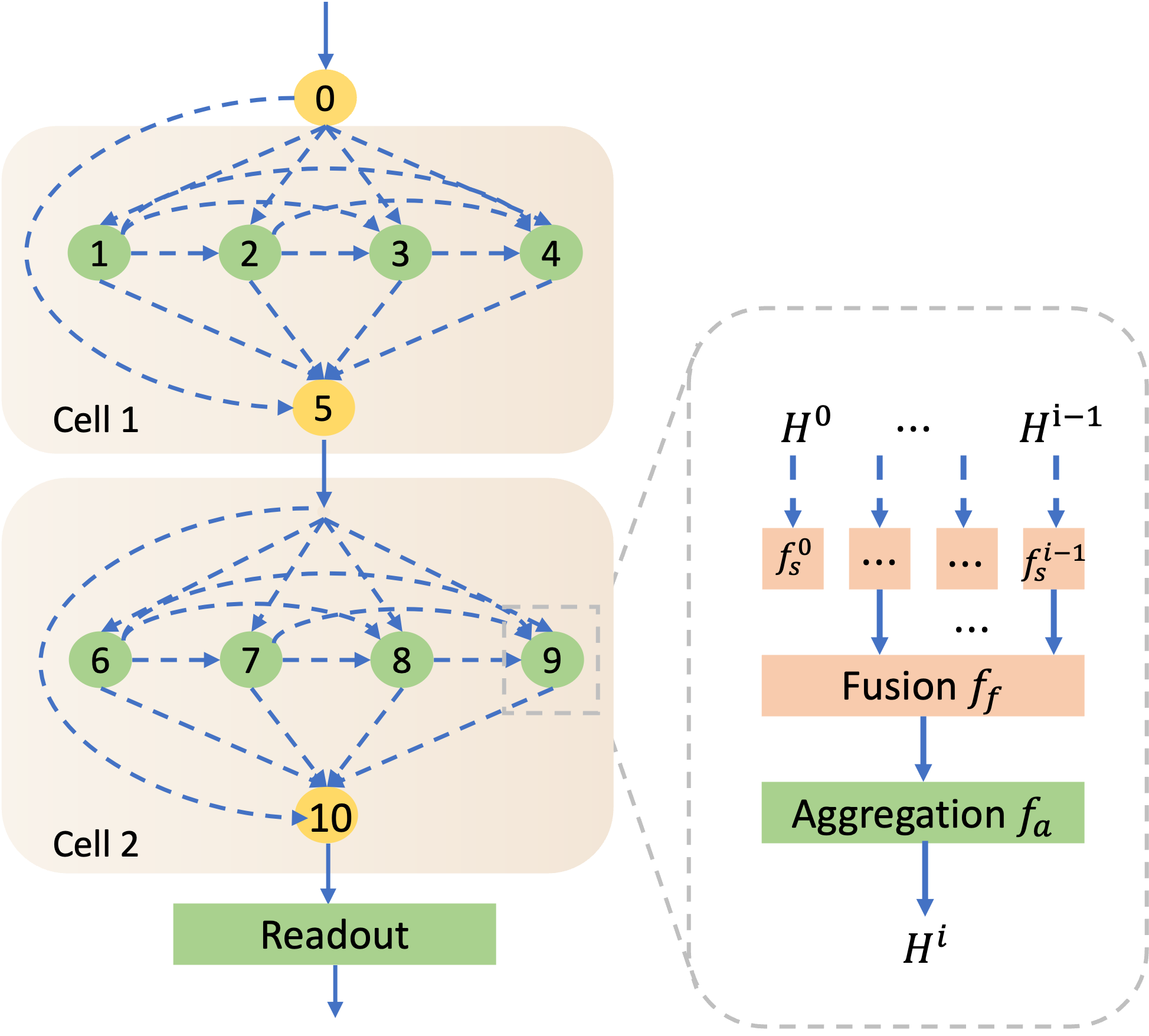}
	\caption{The framework used in LRGNN~\cite{wei2023search}, in which the yellow color nodes represent the pre- and port-processing operations and the green nodes represent the module of selection, fusion and aggregation.}
	\label{fig-lrgnn-framework}
\end{figure}

The graph-level implementation instance is constructed based on method LRGNN~\cite{wei2023search}, which search to capture the long-range dependency with deep stacked GNNs. 
As shown in Fig.~\ref{fig-lrgnn-framework}, the framework are constructed by several cells and in which the selection, fusion and aggregation modules are provided in the same way as shown in Fig.~\ref{fig-backbone}. A readout module is provided to generate the graph representations for this task at the end of framework.

Based on the given framework, we set the basic operations as shown in Table~\ref{tb-space-lrgnn}, on which the LLM-based configuration agents will add or remove operations following the user instructions.

\begin{table}[h]
	\centering
	\footnotesize
	\caption{The candidates of architectures and hyper-parameters used in graph-level tasks.
	}
	\begin{tabular}{c|c|c}
		\toprule
		& Module   & Candidates  \\ \midrule
		\multirow{4}{*}{Model} 
		& Aggregation  & \texttt{GCN}, \texttt{SAGE} \\
		& Selection  & \texttt{ZERO}, \texttt{IDENTITY} \\
		& Fusion  & \texttt{sum}, \texttt{mean} \\ 
		& Readout &global\_sum, Global\_mean \\
		\midrule
		\multirow{4}{*}{\begin{tabular}[c]{@{}c@{}}Hyper-\\ parameters\end{tabular}} 
		& Learning rate   & [0.01, 0.05] \\ 
		& Weight decay & [0.001, 0.005] \\
		& Dropout rate  & [0, 0.5] \\ 
		& Activation & \texttt{ReLU} \\
		\bottomrule
	\end{tabular}
	\label{tb-space-lrgnn}
\end{table}

\subsubsection{Link-level}
\label{sec-appendix-link-instance}
\begin{figure}[ht]
	\centering
	\includegraphics[width=0.8\linewidth]{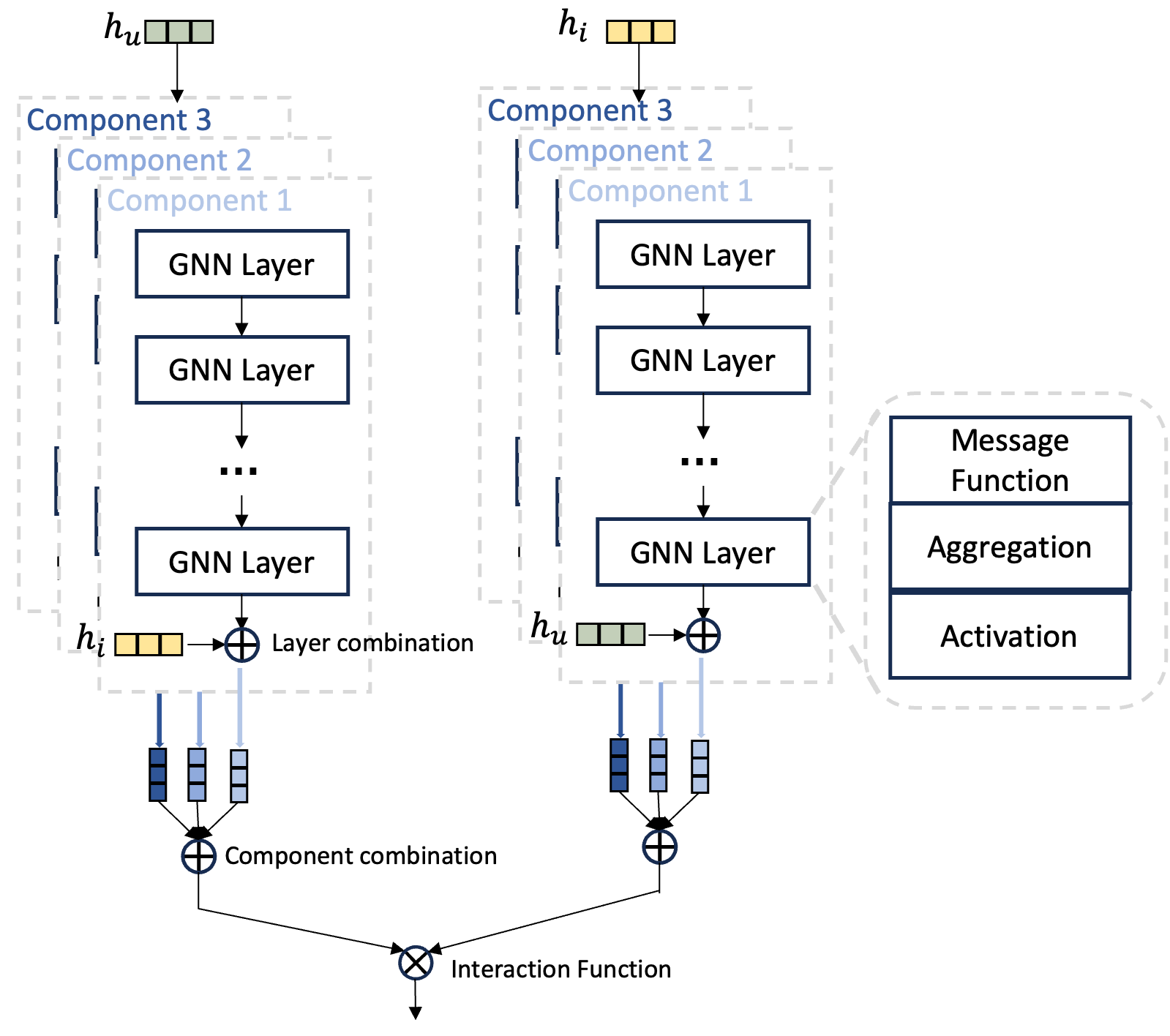}
	\caption{The framework used in Prof-CF \cite{wang2022profiling}.}
	\label{fig-cf-framework}
\end{figure}

For link-level, the instance is constructed based on method Prof-CF~\cite{wang2022automated}, which search for GNN-based two-tower collaborative filtering functions.
As shown in Fig.~\ref{fig-cf-framework}, the framework contains diverse design modules, including message function, aggregation, and activation in each GNN layer, as well as the layer combination, component combination, and interaction functions used beyond layer.

Based on the framework and the constructed search space in Prof-CF, we design a basic candidates as shown in Table~\ref{tb-space- cf}, on top of which the configuration agents will update the candidates following the instructions.
\begin{table}[h]
	\centering
	\footnotesize
	\caption{The candidates of architectures and hyper-parameters used in link-level tasks.
	}
	\begin{tabular}{c|c|c}
		\toprule
		& Module   & Candidates  \\ \midrule
		\multirow{4}{*}{Model} 
		& Message Function & \texttt{Identity}, \texttt{Hadamard} \\ 
		& Aggregation  & \texttt{NONE}, \texttt{GCN}, \texttt{SAGE} \\
		&Layer Number & 1, 3 \\
		& Layer Combination & \texttt{STACK}, \texttt{SUM} \\
		&Component Number & 1,2,3,4 \\
		& Componnet Combination & \texttt{MEAN} \\
		& Interaction Function &\texttt{DOT PRODUCT}, \texttt{CONCAT+MLP} \\ 
		\midrule
		\multirow{4}{*}{\begin{tabular}[c]{@{}c@{}}Hyper-\\ parameters\end{tabular}} 
		& Learning rate   & [0.01, 0.05] \\ 
		& Weight decay & [0.001, 0.005] \\
		& Dropout rate  & [0, 0.5] \\ 
		& Activation & \texttt{ReLU},\texttt{IDENTITY} \\
		\bottomrule
	\end{tabular}
	\label{tb-space- cf}
\end{table}

{\tiny }\subsection{Datasets}
\label{appendix-tb-datasets}
In the node classification task, 
Cora~\cite{sen2008collective} is the citation network where each node represents a paper, and each edge represents the citation relation between two papers; Computers and Photo~\cite{shchur2018pitfalls} are the Amazon co-purchase graphs where nodes represent goods that are linked by an edge if these goods are frequently bought together; Physics~\cite{shchur2018pitfalls} is a co-authorship graph where nodes are authors who are connected by an edge if they co-author a paper. 

The statistics of these datasets are provided in Table~\ref{tb-node-dataset}.
In the graph classification task, NCI1 and NCI109~\cite{wale2006comparison} are datasets of chemical compounds; DD and PROTEINS~\cite{dobsondistinguishing} are datasets both protein graphs. The statistics of these datasets are provided in Table~\ref{tb-graph-dataset}.
In the item ranking task, two datasets are adopted. Epinions
is a graph of consumer reviews, in which node represent the users and the edges represent trust relationship between users. Amazon-Sports is a e-commerce dataset which contains the subset of sports category from~\cite{he2016ups}. The statistics of these datasets are provided in Table~\ref{tb-ir-dataset}.

\begin{table}[ht]
	\setlength\tabcolsep{1pt}
	\centering
	\footnotesize
	\caption{The statistics of four graph classification datasets.}
	\label{tb-graph-dataset}
	\begin{tabular}{c|c|c|c|c|c}
		\toprule
		Dataset & \# Graphs & \# Feature & \# Classes & Avg.\# of Nodes & Avg.\#  of Edges \\ \midrule
		NCI1~\cite{wale2006comparison} & 4,110 & 89 & 2 & 29.87 & 32.3 \\ 
		NCI109~\cite{wale2006comparison} & 4,127 & 38 & 2 & 29.69 & 32.13 \\ 
		DD~\cite{dobsondistinguishing} & 1,178 & 89 & 2 & 384.3 & 715.7 \\ 
		PROTEINS~\cite{dobsondistinguishing} & 1,113 & 3 & 2 & 39.1 & 72.8 \\ 
		\bottomrule
	\end{tabular}
	
\end{table}

\begin{table}[ht]
	\centering
	\caption{The statistics of two datasets used in item ranking task.}
	\begin{tabular}{c|cc}
		\toprule
		& Epinions & Amazon-Sports \\ \midrule
		\# Users & 40,163 & 11,435 \\
		\# Items & 139,738 & 5, 405 \\
		\# Interactions & 664, 824 & 108, 004 \\
		\# Rating Scale & {[}1,5{]} & {[}1, 5{]} \\ \midrule
		Density & 0.012\% & 0.17\% \\ \bottomrule
	\end{tabular}
	\label{tb-ir-dataset}
\end{table}

\begin{table}[ht]
	\centering
	\footnotesize
	\caption{The statistics of four node classification datasets.}
	\begin{tabular}{c|c|c|c|c}
		\toprule
		Datasets                          & \#Nodes & \#Edges & \#Features & \#Classes \\ \midrule
		Cora~\cite{sen2008collective}     & 2,708   & 5,278   & 1,433      & 7        \\ 
		Computers~\cite{shchur2018pitfalls} & 13,381  & 245,778 & 767        & 10        \\
		Photo~\cite{shchur2018pitfalls} & 7,487 & 238,162 & 745 &8  \\
		Physics~\cite{shchur2018pitfalls} & 34,493  & 495,924 & 8,415      & 5       \\ 
		genius~\cite{lim2021new} & 421,961 &984, 979 &12 &2 \\
		\bottomrule
	\end{tabular}
	\label{tb-node-dataset}
\end{table}

\subsection{Baselines}
\label{sec-appendix-llmgnn}
We use three types of baselines in this paper to evaluate the versatility of GL-Agent, i.e., the human-designed widely used baselines and SOTA methods used in recent two years; the AutoML-based method; and the baselines that suggested by LLMs directly. In the following, we introduce the baselines used in each task, and then analyze the construction of LLL-GNN methods in each dataset.

\noindent\textbf{Node-level}. We adopt the (1) human-designed method. Four-layer GCN and GIN baseline; GPR-GNN~\cite{chien2020adaptive} that learns the weights of each layer based on generalized PageRank, and ACM-GCN~\cite{luan2022revisiting} that adaptive mixing the channel information from low-/high-/full-frequency, and the configuration of these two methods are followed the original paper.
(2) NAS-based method SANE~\cite{zhao2021search} that learns the connections based on JKNet~\cite{xu2018representation} and F2GNN~\cite{wei2022designing} that designs the network topology from the feature fusion perspective. We employ the official code and then search on the target dataset.

\noindent\textbf{Graph-level}. We adopt the (1) the human-designed global pooling method. Four-layer GCN and GIN baseline with global add readout operations; DGCNN~\cite{zhang2018end} that learn the graph representation based on the selected top-ranked nodes;
and the hierarchical pooling methods DiffPool~\cite{ying2018hierarchical} to learn the hierarchy nature in the graphs; (2) NAS-based method. LRGNN~\cite{wei2023search} that design the network topology to capture the long-range dependency. The experiments are conducted based on the official code provided by this paper.

\noindent\textbf{Link-level}. In the item ranking task, we adopt (1) the human-designed method. NCF~\cite{he2017neural} leverages neural networks to learn a more expressive interaction functions; NGCF~\cite{wang2019neural} and LightGCN~\cite{he2020lightgcn} are the representative GNN-based collaborative filtering methods, where the former adopts hadamard product and the latter uses identity function in the message function. (2) NAS-based method. Prof-CF~\cite{wang2022profiling} aim to design an expressive search space by pruning operations. We  conduct the experiments following the settings and using the codes provided by this paper.

\noindent\textbf{The construction of LLM-GNN baseline}.
Apart from that , we provide the LLM-GNN baseline, for which the GNN and hyper-parameters are suggested by LLM (GPT-3.5-turbo) given
given the dataset description and statistics, the details can be found in Table~\ref{tb-llm-gnn}.  They are obtained based on the following prompts:

\texttt{Your are an expert on graph learning. Firstly, could you please describe the dataset \{$\mathcal{D}$\} used in the task \{$\mathcal{T}$\}. 
	Then, you can suggest one GNN that could achieve better performance on this dataset.
	Here are the design dimensions you can refer to: the aggregation operation (message-passing layer), the activation function, the layer numbers of the designed GNN,
	the skip-connections beyond layer, the function that integrating the features from different layers, the hyper-parameters, and etc.}

For the non-homophilous dataset genius, we further provide the dataset introduction and experimental observations as shown in the following:

\texttt{This is a non-homophilous dataset that connected nodes may have different labels. For this dataset, using MLPs may have better performance than general GNNs.}

For the item ranking task, we provide one additional sentence in the prompt:

\texttt{Multiple GNNs can be added when extracting the results, and then you can choose the numbers of GNNs and the combination function.}

\begin{table}[ht]
	\scriptsize
	\centering
	\caption{The suggested solutions used in LLM-GNN baseline.}
	\begin{tabular}{l|L{6cm}}
		\toprule
		Data & Suggestions \\ \midrule
		Cora & GNN: Two-layer stakcing-based GCN. Hyper-parameters: \{ hidden size:16, dropout ratio:0, learning rate:0.01, weight decay:5e-4 \}    \\ \midrule
		Computer & GNN: Two-layer stakcing-based GCN. Hyper-parameters: \{ 'hidden':256, 'dropout':0.1, 'lr':0.01, 'l2':5e-4,' \} \\ \midrule
		Photo & GNN: Two-layer GCN in which the aggregation results are concatenated to predict the node labels.  Hyper-parameters: \{ 'hidden':128, 'dropout':0.1, 'lr':0.005, 'l2':5e-4,  \} \\ \midrule
		Physics & GNN: Two-layer GCN in which the aggregation results are concatenated to predict the node labels. Hyper-parameters: \{ 'hidden':128, 'dropout':0.1, 'lr':0.005, 'l2':5e-4,  \} \\  \midrule
		genius & GNN: Two-layer GraphSAGE with mean aggregator. Hyper-parameters: \{`activation': ReLU, hidden size: 128, lr: 0.01, 'l2':5e-4\}. \\ \midrule
		DD & GNN: Three-layer GIN in which the aggregation results are added, and graph representation vector is obtained with Global sum readout operation. Hyper-parameters: \{ 'hidden':128, 'dropout':0.1, 'lr':0.01, 'l2':5e-4 \}. \\ \midrule
		PROTEINS & GNN: Two-layer stakcing-based GIN, and graph representation vector is obtained with Global sum readout operation. Hyper-parameters: \{ 'hidden':32, 'dropout':0, 'lr':0.01, 'l2':5e-4 \}. \\ \midrule
		NCI1 & GNN: Three-layer GIN in which the aggregation results are added, and graph representation vector is obtained with Global sum readout operation. Hyper-parameters: \{ 'hidden':64, 'dropout':0.0, 'lr':0.01, 'l2':5e-4 \}. \\ \midrule
		NCI109 & GNN: Three-layer GIN in which the aggregation results are added, and graph representation vector is obtained with Global sum readout operation. Hyper-parameters: \{ 'hidden':128, 'dropout':0.1, 'lr':0.01, 'l2':5e-4 \} \\ \midrule
		Epsonion & Three components are used in the GNN. In each component, two-layer stakcing-based GCN are employed, and these components are concatenated to obtain the node representations. The user and item rep- resentations are first concatenated or summed up, and then fed into an MLP for prediction \\ \midrule
		Amazon-Sports & GNN: Two-layer GAT in which the aggregation results are concatenated to formulate the final node representations. The Hadamard operation is adopted when calculating the messages. The final prediction is obtained based on the dot product of the user and item representations. 
		\\ \bottomrule
	\end{tabular}
\label{tb-llm-gnn}
\end{table}

\subsection{Performance Comparisons}
\label{sec-appendix-perf}
In this section, we show the performance comparisons with the baselines to evaluate the effectiveness of the proposed method and the implemented instance in graph learning over different tasks. 
As indicated in Table~\ref{tb-perf-node-full}, our method outperforms all others across four node classification datasets, even when compared to NAS-based methods that focus on either architecture topology or aggregation operation.
For LLM-GNN, where the GNN and hyper-parameters are directly suggested by LLM, the performance is subpar compared to the GCN baseline due to inappropriate hyper-parameters. They fail to outperform well-designed baselines provided by research, such as GPRGNN and ACM-GNN used in this table. This highlights the challenge LLMs face in keeping up-to-date with the latest knowledge in graph learning. The suboptimal performance of the suggested GCN baseline further supports this conclusion. In contrast, our method, which designs the flexible and versatile graph learning approach with the assistance of LLM-based agents and AutoML, rather than solely relying on LLMs, exhibits superior performance.
Similar results are observed in the graph classification datasets in Table~\ref{tb-perf-graph-full}. LLM-GNN struggles to outperform existing methods, whether they overlook hierarchical information or use models with insufficient depth~\cite{wei2023search,wei2021pooling}.

For the item ranking task, we evaluate the Recall@20 performance on two datasets as shown in Table~\ref{tb-perf-ir}. 
It can be observed that GL-Agent achieves higher performance than human-designed methods, which demonstrates its effectiveness.
When compared to Prod-CF~\cite{wang2022profiling}, which GL-Agent is based on, it neglects multiple components in GNNs as shown in Table~\ref{tb-prompt-all-tasks}, resulting in inferior performance. Furthermore, when we use LLM to directly suggest GNNs, we mention multiple components in the prompt, but only Epinions considers this design dimension when designing GNNs. These findings empirically demonstrate the outdated information maintained in LLMs and their limited ability to design more powerful GNNs based on the latest prior knowledge.

The evaluations conducted on three tasks demonstrate the feasibility of GL-Agent in managing diverse user requirements that may arise in real-world scenarios. It has the potential to achieve superior or comparable performance to human-designed methods, using only instructions. This makes it a flexible and user-friendly tool. Given its low requirements for coding ability and domain knowledge of graph learning, GL-Agent is particularly accessible to non-expert users, including those unfamiliar with graph learning. From this perspective, it serves as a versatile tool for designing effective graph learning solutions.

\begin{table}[htb]
	\caption{The performance comparisons on the node classification task.}
	\footnotesize
	\centering
	\begin{tabular}{c|cccc|c}
			\toprule
			& Cora        & Photo       & Computer    & Physics     & Avg. Rank \\ \midrule
			GL-Agent & 86.81(0.40) & 96.40(0.16) & 92.47(0.20) & 96.93(0.98) & 2.25      \\ \midrule
			GCN            & 85.68(0.61) & 93.13(0.27) & 90.52(0.42) & 95.97(0.14) & 5.5       \\ 
			GIN            & 83.83(1.36) & 92.67(0.57) & 88.67(1.21) & 95.79(0.16) & 7.5       \\  \midrule
			GPRGNN         & 87.62(0.48) & 91.93(0.26) & 88.90(0.37) & 97.51(0.21) & 4.5       \\  
			ACM-GCN        & 86.67(0.14) & 94.35(0.65) & 88.58(0.39) & 97.89(0.23) & 4.5       \\  \midrule
			SANE           & 86.40(0.38) & 94.53(0.22) & 90.25(0.31) & 98.28(0.13) & 3.25      \\  
			F2GNN          & 87.42(0.42) & 95.38(0.30) & 91.42(0.26) & 96.92(0.06) & 2.75      \\ \midrule
			LLM-GNN        & 84.64(1.04) & 93.73(0.38) & 89.20(1.16) & 96.34(0.11) & 5.75      \\ \bottomrule
		\end{tabular}
	\label{tb-perf-node-full}
\end{table}

\begin{table}[htb]
	\footnotesize
	\centering
	\caption{The performance comparisons on the graph classification task.}
	\begin{tabular}{c|cccc|c}
			\toprule
			& Proteins    & DD          & NCI1        & NCI109      & Avg. Rank \\ \midrule
			GL-Agent & 75.38(5.03) & 78.10(3.21) & 82.14(1.74) & 81.25(1.53) & 2       \\ \midrule
			GCN            & 74.84(3.07) & 73.59(4.17) & 76.96(3.07) & 75.70(4.03) & 4.25      \\ 
			GIN            & 74.50(4.10) & 74.62(2.74) & 77.95(1.95) & 73.25(2.67) & 4.5       \\ \midrule
			DGCNN          & 73.95(3.04) & 61.63(5.33) & 76.08(1.03) & 74.58(3.99) & 5.75      \\
			DiffPool       & 75.11(2.14) & 77.85(3.53) & 75.04(1.98) & 71.48(2.46) & 4.75      \\  \midrule
			LRGNN          & 75.39(4.40) & 78.18(2.02) & 82.51(1.37) & 81.39(1.92) & 1       \\  \midrule
			LLM-GNN        & 74.47(3.65) & 75.12(3.44) & 71.70(2.58) & 73.04(3.65) & 5.75      \\ \bottomrule
		\end{tabular}
\label{tb-perf-graph-full}
\end{table}


\subsection{Case Study}
\label{sec-case-show}
\begin{figure*}[ht]
     \includegraphics[width=23cm, height=14.375cm, angle=270]{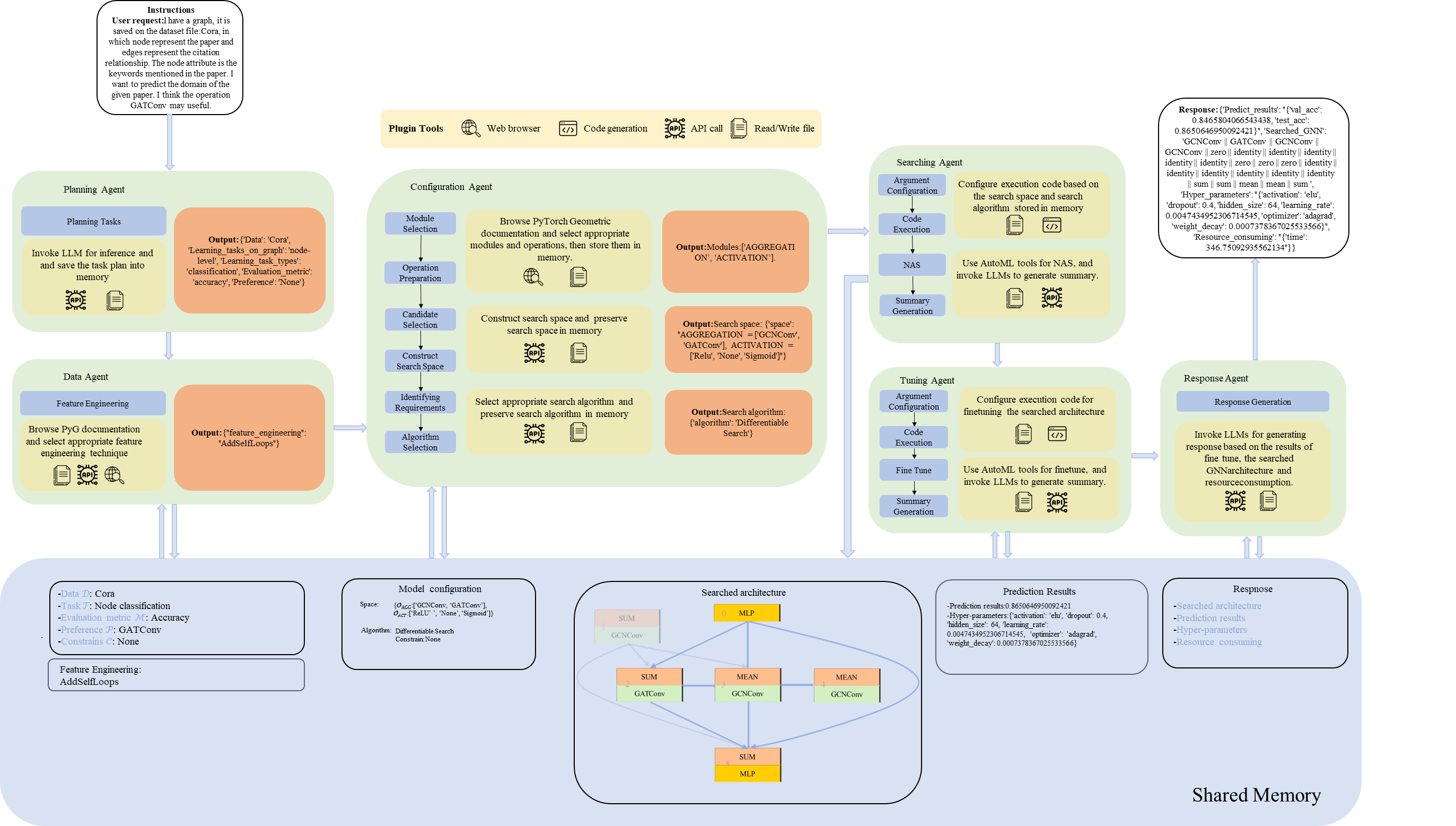}
    \caption{Case Study on Cora Dataset.}
    \label{case-study-node-level}
\end{figure*}

\begin{figure*}[ht]
     \includegraphics[width=23cm, height=14.375cm, angle=270]{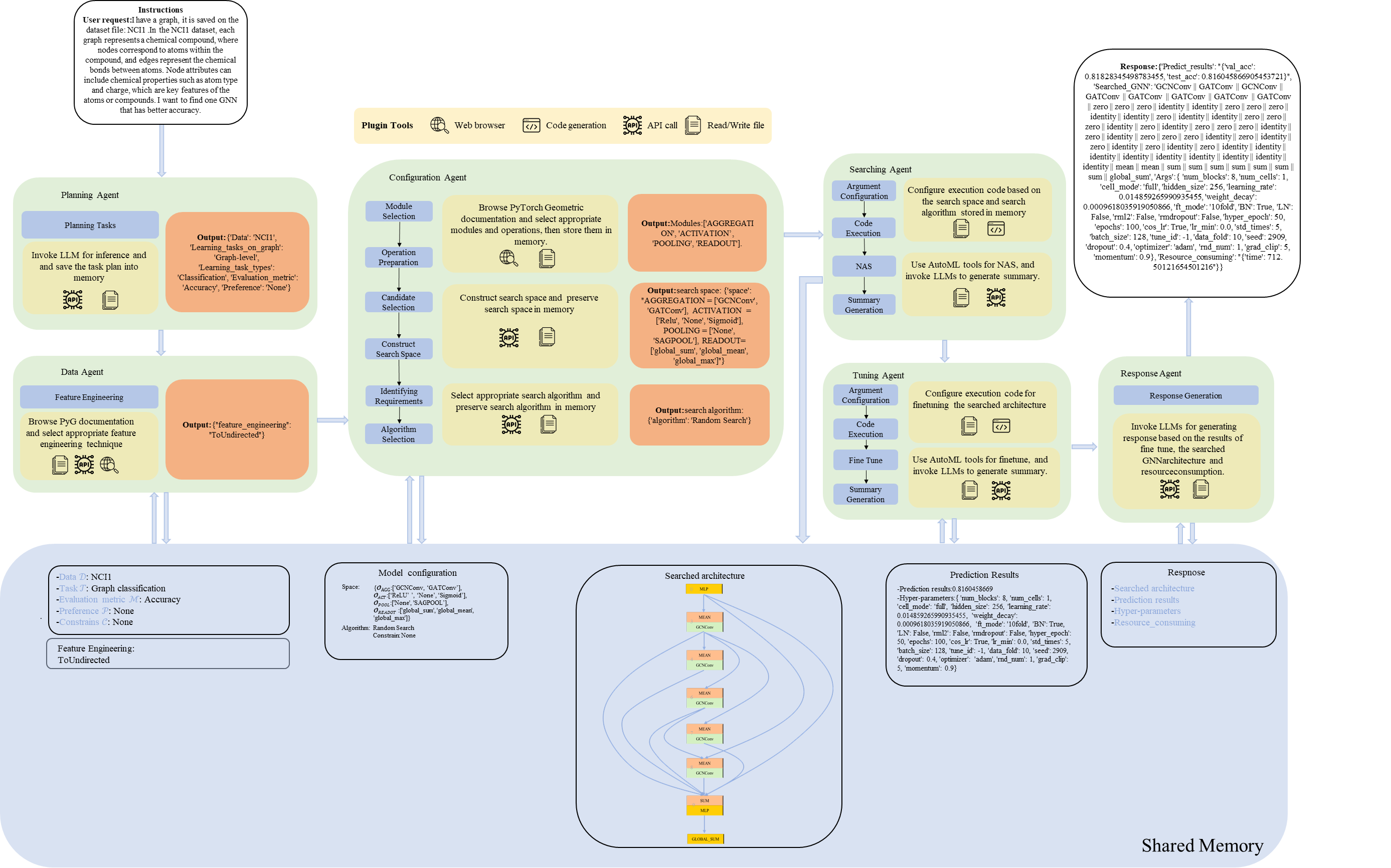}
    \caption{Case Study on NCI1 Dataset}
    \label{case-study-graph-level}
\end{figure*}

In this section, we will present specific case studies to demonstrate the feasibility and versatility of the GL-Agent framework. Fig.~\ref{case-study-node-level} and Fig.~\ref{case-study-graph-level} illustrate the entire process, from the specific request posed by the user to the final results of searched GNN models.  In these figures, we clearly observe that the correctness of decisions and execution.

%

\subsection{Interpretability of Decisions in Graph Learning Tasks}
\label{sec-appendix-Interpretability}
User requirements may involve data structures and diverse analysis goals. The interpretability of the decisions made by LLM-based agents is self-evidently important to ensure that users can clearly understand the model recommendations. 
Therefore, we explore in detail the thinking of LLM when parsing input and formulating strategies accordingly. 
Table~\ref{tb-agent-making-decision} shows the decision-making process of each agent when handling different tasks.

\clearpage
\begin{table*}[htb]
	\scriptsize
	\caption{The various user instructions.}
	\begin{tabular}{L{0.8cm}|L{4.5cm}|L{5cm}|L{5cm}}
		\toprule
		Number & Original User Instruction & User Instruction Paraphrased by ChatGPT-4 & User Instruction Misled by ChatGPT-4\\ \midrule
		
		1 
		& I have a graph, it is saved on the dataset file: Cora, in which node represent the paper and edges represent the citation relationship. The node attribute is the keywords mentioned in the paper.I want to predict the domain of the given paper. 
		& I'm working with a graph dataset called Cora. In this dataset, each node corresponds to an academic paper, while the edges between nodes signify citation relationships among the papers. Additionally, each node is associated with attributes that capture the keywords mentioned in the respective papers. My goal is to leverage this dataset to predict the academic domain or field of study to which a given paper belongs.
		& I have a graph, it is stored in the dataset file: Cora, where each node supposedly represents a scientific study, and edges falsely suggest a co-authorship relation. The node attribute misleadingly includes the main topics discussed in the study. My goal is erroneously stated as predicting the academic field of the corresponding study based on these misrepresented connections.
		\\ \midrule

		2 
		& I have a graph, it is saved on the dataset file: Computer. This dataset is based on the 'Computers' section of the Amazon co-purchase graph. In this graph, nodes represent products, and edges connect products that are frequently bought together. I am looking to leverage this graph to build a recommendation system that suggests products based on co-purchase history. 
		& I'm working with a graph dataset named "Computer", derived from the "Computers" section of the Amazon co-purchase graph. In this dataset, each node represents a product, and edges between nodes indicate products that are frequently purchased together. My objective is to utilize this graph to develop a recommendation system that suggests products based on their co-purchase history.
		& I possess a graph housed in the dataset named 'Computer,' extracted from the Amazon co-purchase data focused on electronics. Within this structure, each node misleadingly symbolizes a consumer, and the edges wrongly imply financial transactions between them. My aim is to develop a system that predicts future buying trends of these consumers based on their supposed transaction history.
		\\ \midrule

		3 
		& I have a graph, it is saved on the dataset file: physics. It shows authors as nodes, and if two authors wrote a paper together, they're connected by an edge. I want to use this to find groups of authors who work closely and guess who might team up on future projects. 
		& I'm working with a graph dataset named "physics", where nodes represent authors, and edges connect authors who have collaborated on writing a paper together. My aim is to analyze this graph to identify clusters of authors who collaborate closely and predict potential future collaborations among them.
		& I have a dataset stored in the 'physics' file, where each node is mistakenly thought to represent a scientific publication, and the edges erroneously indicate that these publications are based on similar research topics. My goal is to predict future collaborative research topics among these publications, assuming they might combine their findings.
		\\ \midrule

		4 
		& I have a graph, it is saved on the dataset file: actor, where each point is an actor, and a line between two actors means they're mentioned together on a Wikipedia page. I want to look at how these actors are connected, find groups, and see how they relate to each other. 
		& I'm analyzing a graph dataset named "actor", wherein each node represents an actor, and an edge between two actors indicates that they are mentioned together on a Wikipedia page. My objective is to examine the connections between these actors, identify groups or clusters within the network, and explore the relationships between them.
		& I have accessed a graph data file named 'actor', where each point misleadingly represents a movie, and the lines between points incorrectly imply that actors from these movies have worked together on the same film set. My objective is to analyze the movie genres to find potential future collaborations among these actors based on their genre appearances.
		\\ \midrule

		5 
		& I have a graph, it is saved on the dataset file: wisconsin. This graph is a map of web pages, where each node is a page and each edge is a link to another page.The pages have features based on the content and hyperlinks. My task is to categorize the pages. 
		& I am working with a dataset titled 'Wisconsin,' which consists of a graph stored at /data/wisconsin/. In this graph, each node corresponds to a web page, and each edge represents a hyperlink that connects one page to another. Attributes of each page include content details and hyperlink information. My task is to categorize these web pages into distinct classes based on their attributes and the structure of their hyperlinks.
		& I have a dataset called 'wisconsin' that is inaccurately described as a geographic map of physical locations, with each node representing a location and each edge depicting roads connecting them. The features of these nodes supposedly include demographic data and transport links. My goal is to classify these locations based on their demographic characteristics.
		\\ \midrule

		6 
		& I have a graph, it is saved on the dataset file: NCI1. In the NCI1 dataset, each graph represents a chemical compound, where nodes correspond to atoms within the compound, and edges represent the chemical bonds between atoms. Node attributes can include chemical properties such as atom type and charge, which are key features of the atoms or compounds. I want to find one GNN that has better accuracy.
		& I'm working with a graph dataset named "NCI1". In this dataset, each graph represents a chemical compound, with nodes corresponding to atoms within the compound and edges representing the chemical bonds between atoms. Additionally, node attributes capture important chemical properties such as atom type and charge, which are crucial features for characterizing the atoms or compounds. My goal is to identify a Graph Neural Network (GNN) model that achieves higher accuracy in analyzing this chemical compound dataset.
		& I have a graph dataset named 'NCI1'. Each graph in this dataset represents a chemical compound, with nodes corresponding to atoms and edges representing the chemical bonds between them. Node attributes include important chemical properties like atom type and charge, which are essential for characterizing the atoms or compounds. I aim to identify a Graph Neural Network (GNN) model that offers superior accuracy in analyzing these chemical structures.
		\\ \midrule

		7 
		& I have a graph, it is saved on the dataset file: Epinions, in which nodes represent the users and the edges represent the trust relationship between users. The node attribute could include user activity metrics, ratings, or other relevant information that signifies the user's influence or trustworthiness. I want to predict the potential of a trust relationship forming between two users. 
		& I'm analyzing a graph dataset named "Epinions", where nodes represent users and edges represent the trust relationships between users. Node attributes may include user activity metrics, ratings, or other relevant information indicating the user's influence or trustworthiness. My objective is to develop a predictive model that can anticipate the likelihood of a trust relationship forming between two users based on the available data in the graph.
		& I am working with a graph dataset named 'Epinions', stored in a specific dataset file. In this graph, each node represents a user, and the edges denote the trust relationships existing between users. The node attributes might include various user activity metrics, ratings, or other pertinent information that highlights a user’s influence or trustworthiness. My objective is to develop a predictive model that assesses the potential for forming a trust relationship between two users.
		\\ \midrule

		8 
		& In a movie recommendation system, the data is stored at the path: /data/movies/. Within this system, user nodes represent users, and edges represent their social connections. I want to predict users' movie preferences, i.e., which types of movies users are likely to enjoy. 
		& Within our movie recommendation system, the dataset is located at /data/movies/. In this system, user nodes represent individual users, while edges symbolize their social connections. My goal is to predict users’ movie preferences, specifically identifying the types of movies users are likely to enjoy.
		& In our movie recommendation system, the dataset is stored at the location /data/movies/. In this system, each node within the graph represents a user, while the edges illustrate the social connections between these users. The goal is to predict the movie preferences of these users, specifically determining which types of movies are likely to appeal to each user based on their social connections.
		\\ \midrule

		9 
		& In a protein-protein interaction network, the dataset is located at /data/proteins/. Nodes represent proteins and edges represent interactions between them. I'm interested in classifying proteins based on their functions. The node attributes include the type of protein and its biological properties. I believe GCNConv might be a good fit for this task. 
		& In the protein-protein interaction network, the dataset is stored at /data/proteins/. Nodes correspond to proteins, while edges represent interactions between them. My objective is to classify proteins based on their functions. The node attributes include the type of protein and its biological properties. I believe utilizing the GCNConv (Graph Convolutional Network Convolution) method might be suitable for achieving this task.
		& In the protein-protein interaction network housed at /data/proteins/, each node is a protein and each edge signifies an interaction between proteins. My focus is on classifying these proteins based on their functions, utilizing the node attributes that detail each protein’s type and biological properties. I am considering using the Graph Convolutional Network (GCNConv) model, as it might be well-suited for analyzing the complex relationships and properties encapsulated in this dataset.
		\\ \midrule

		10 
		& For a chemical reaction network located at /data/chemical\_reactions/, nodes are reactants/products and edges are reaction pathways. The goal is to predict reaction outcomes.
		& In the chemical reaction network stored at /data/chemical\_reactions/, nodes represent reactants and products, while edges denote reaction pathways. The objective is to predict reaction outcomes.
		& In the chemical reaction network stored at /data/chemical\_reactions/, each node represents either a reactant or a product, and the edges delineate the pathways of chemical reactions between these nodes. The objective of this study is to predict the outcomes of these reactions by analyzing how reactants transform into products along the defined pathways. This analysis aims to enhance our understanding of chemical reaction dynamics and potentially predict new reaction outcomes based on existing data.
		\\ \bottomrule
	\end{tabular}
	\label{tb-user-request}
\end{table*}

\clearpage
\begin{table*}[htb]
	\centering
	\scriptsize
	\caption{Interpretation of Agents Decisions in Diverse Graph Learning Scenarios }
	\begin{tabular}{L{2cm}|L{4cm}|L{6cm}|L{4cm}}
		\hline
		Agents & Input information & Reason of decision & Agent output \\ \hline                                                                                                                                                                                                                                                                                                                                                                                                                                                               
		Manager agent
		& \textbf{User Request :} I have a graph, it is saved on the path:Cora, in which node represent the paper and edges represent the citation relationship. The node attribute is the keywords mentioned in the paper.I want to predict the domain of the given paper.
		& \{'Learning\_tasks\_on\_graph\_reason': 'The user wants to predict the domain of the given paper. The node attribute is the keywords mentioned in the paper. Therefore, the node-level learning task is chosen.', 'Learning\_task\_types\_reason': 'The user wants to predict the domain of the given paper. Therefore, the classification learning task is chosen.', 'Evaluation\_metric\_reason': 'The user wants to predict the domain of the given paper. Therefore, the accuracy metric is chosen.'\}
		& \{'Data': 'Cora', 'Learning\_tasks\_on\_graph': 'node-level', 'Learning\_task\_types': 'classification', 'Evaluation\_metric': 'accuracy', 'Preference': 'None'\} \\ \hline
		
		Manager agent
		& \textbf{User Request :} In an environmental monitoring network at /data/environment\_network/, nodes are sensor locations and edges represent ecological connections. The goal is to predict the overall health of an ecosystem. Graph-level attributes include biodiversity indices and pollution levels.
		& \{'Learning\_tasks\_on\_graph\_reason': 'The goal is to predict the overall health of an ecosystem, which is a graph-level attribute. Therefore, a graph-level learning task is chosen.', 'Learning\_task\_types\_reason': 'The goal is to predict a continuous variable (overall health), which is a regression task.', 'Evaluation\_metric\_reason': 'R-squared is a common metric for regression tasks, and it is preferred because it provides a measure of how well the model fits the data.'\}
		& \{'Data': '/data/environment\_network/', 'Learning\_tasks\_on\_graph': 'graph-level', 'Learning\_task\_types': 'regression', 'Evaluation\_metric': 'R-squared', 'Preference': 'None'\} \\ \hline
		
		Manager agent
		& \textbf{User Request :} In a book co-authorship network stored at /data/book\_network/, nodes are authors and edges are co-authorships. The goal is to identify influential authors in different genres. I'm leaning towards using GCNConv.
		& \{'Learning\_tasks\_on\_graph\_reason': 'The goal is to identify influential authors in different genres. ', 'Learning\_task\_types\_reason': 'The user wants to identify influential authors in different genres. Classification is used.', 'Evaluation\_metric\_reason': 'Accuracy is used as the evaluation metric because the user wants to identify influential authors in different genres.'\}
		& \{'Data': '/data/book\_network/', 'Learning\_tasks\_on\_graph': 'node-level', 'Learning\_task\_types': 'classification', 'Evaluation\_metric': 'accuracy', 'Preference': ' GCNConv '\} \\ \hline
		
		Data agent 
		& \textbf{User Request :} I have a graph, it is saved on the dataset file:Cora, in which node represent the paper and edges represent the citation relationship.The node attribute is the keywords mentioned in the paper.I want to predict the domain of the given paper. \textbf{Task plan :} \{'Data': 'Cora', 'Learning\_tasks\_on\_graph': 'node-level', 'Learning\_task\_types': 'classification', 'Evaluation\_metric': 'accuracy', 'Preference': 'None'\}
		& \{"feature\_engineering\_reason": "The AddSelfLoops transform is used to add self-loops to the given homogeneous or heterogeneous graph. This is useful for cases where the graph does not have self-loops, but the model requires them to be present."\}
		& \{"feature\_engineering": "AddSelfLoops"\}\\ \hline

		Data agent 
		& \textbf{User Request :} In an online gaming network at /data/gaming\_network/, nodes are players and edges are interactions. The goal is to identify user engagement patterns. Node attributes include playtime and activity data. 
		\newline \textbf{Task plan :} \{'Data': '/data/gaming\_network/', 'Learning\_tasks\_on\_graph': 'Node-level', 'Learning\_task\_types': 'Classification', 'Evaluation\_metric': 'Accuracy', 'Preference': 'None'\}
		& \{"feature\_engineering\_reason": "The gaming network is undirected, hence the ToUndirected feature engineering technique is suitable."\}
		& \{feature\_engineering": "ToUndirected"\} \\ \hline

		Configuration agent
		& \textbf{Task plan :} \{'Data': 'Cora', 'Learning\_tasks\_on\_graph': 'Node-level', 'Learning\_task\_types': 'Classification', 'Evaluation\_metric': 'Accuracy', 'Preference': 'None'\}
		& \{'reason': 'The task is related to graph representation learning, and the hidden size cannot adopt the differentiable search algorithm. The design dimension does not contain any coarsening operations. Therefore, the differentiable search algorithm is suitable for this task.'\}
		& \{'algorithm': 'Differentiable Search'\} \\ \hline
		
	\end{tabular}
	\label{tb-agent-making-decision}
\end{table*}

\begin{table*}[htb]
	\centering
	\scriptsize
	\caption{The comparisons of agent results based on different LLMs. }
	\begin{tabular}{l|L{2cm}|L{2cm}|L{2cm}|L{2cm}|L{2cm}}
			\hline
			&        & LLaMA2-7B & LLaMA2 70B                                                                                                                                                                         & GPT-3.5-turbo + LangChain (ours)                                                                          & GPT-4                                                                                                                                                                                                                                                                                                      \\ \hline
			\multirow{3}{*}{Configuration} 
			& Module Selection     
			&   Possible operations:  Aggregation: $\cdots$, Pooling: $\cdots$, Readout: $\cdots$ 
			& Here are my suggestions for the three operations provided, along with their justifications, for the graph-level task: Convolution: $\cdots$, Pooling: $\cdots$, Readout: $\cdots$ 
			& The response list is [`aggregation', `pooling', `readout',`selection',`fusion'] 
			& The graph-level task involves learning a representation for the entire graph, which is often used for graph classification or regression. Convolution: Yes. $\cdots$, Pooling: Yes. $\cdots$, Readout: Yes. $\cdots$ .So, all the operations [convolution, pooling, readout] can be used for graph-level tasks. \\ \cline{2-6} 
			& Operation Preparation \& Candidate Selection
			&  \gray To address your request: 1. Justification: $\cdots$, 2. Finding the corresponding class: $\cdots$, 3. Returning the class name and module name $\cdots$.
			& The class name for the convolution operation would be ChebConv. The module name would be pyg.nn.                                                                                  
			& 'ChebConv',  'torch\_geometric.nn.conv.cheb \_conv.ChebConv'                     
			& This function is part of the convolution module in PyTorch Geometric. So, the class name is ChebConv and the module name is convolution.                                                                                                                                                                        \\ \cline{2-6} 
			& Formatted output     
			&   \{ 'aggregation': ['GCNConv', 'GATConv', 'ChebConv'], "READOUT":$\cdots$ \}
			& \{ 'aggregation': ['GCNConv', 'GATConv', 'ChebConv'], "READOUT":$\cdots$ \}                                                                                                         
			& \{`aggregation': [ GCN, SAGE, ChebConv ],`readout': $\cdots$\}                  
			& \{ 'aggregation': ['GCNConv', 'GATConv', 'ChebConv'], "READOUT":$\cdots$ \}                                                                                                                                                                                                                                       \\ \hline
			Algorithm                      
			& Identifying Requirements \& Algorithm Selection 
			&  \gray Firstly, let me clarify the difference between them. $\cdots$. Then, Lets evaluate the options based on these principles. $\cdots$     
			& Recommendation: Differentiable Search Algorithm                                                                                                                                  
			& You should use``differentiable search algorithm".                               
			& $\cdots$ So, in this case, both differentiable and random search algorithms should be used depending on the specific operation.                                                                                                                                                                                 \\ \hline
		\end{tabular}
	\vspace{-5pt}
	\label{tb-comparisons-llms}
\end{table*}



\end{document}